\documentclass[10pt,logo,copyright]{nvidiatechreport}
\usepackage[authoryear,round]{natbib}

\usepackage[utf8]{inputenc} 
\usepackage[T1]{fontenc}    

\usepackage{parskip}        
\usepackage{url}            
\usepackage{booktabs}       
\usepackage{amsfonts}       
\usepackage{nicefrac}       
\usepackage{microtype}      
\usepackage{xcolor}         
\usepackage[dvipsnames]{xcolor} 
\usepackage{graphicx}
\usepackage{animate}        
\usepackage{subcaption}
\usepackage{tabularx}
\usepackage{makecell}
\usepackage{adjustbox}
\usepackage{setspace}
\newcolumntype{M}[1]{>{\centering\arraybackslash}m{#1}}
\usepackage{float}
\usepackage{tikz}
\usetikzlibrary{positioning,shapes,shapes.geometric,arrows, decorations.pathreplacing, backgrounds}
\usepackage{amsmath,amsfonts,bm, bbm}
\usepackage{multirow}
\usepackage{comment}
\usepackage{gensymb}
\usepackage{lipsum}
\usetikzlibrary{arrows.meta, positioning, fit}
\usepackage[para]{threeparttable}
\usetikzlibrary{tikzmark}

\usepackage[most]{tcolorbox}
\usepackage{fancyvrb}
\usepackage{fvextra}
\usepackage{dashrule}

\usepackage{multicol}

\makeatletter
\renewcommand{\paragraph}{%
  \@startsection{paragraph}{4}{\z@}%
    {0.5ex \@plus .2ex \@minus .2ex}
    {-0.4em}
    {\normalfont\normalsize\bfseries}%
}
\makeatother

\def\eqref#1{equation~\ref{#1}}

\def\1{\bm{1}}

\DeclareMathAlphabet{\mathsfit}{\encodingdefault}{\sfdefault}{m}{sl}
\SetMathAlphabet{\mathsfit}{bold}{\encodingdefault}{\sfdefault}{bx}{n}

\makeatletter
\let\save@mathaccent\mathaccent
\newcommand*\if@single[3]{%
  \setbox0\hbox{${\mathaccent"0362{#1}}^H$}%
  \setbox2\hbox{${\mathaccent"0362{\kern0pt#1}}^H$}%
  \ifdim\ht0=\ht2 #3\else #2\fi
  }
\newcommand*\rel@kern[1]{\kern#1\dimexpr\macc@kerna}
\newcommand*\widebar[1]{\@ifnextchar^{{\wide@bar{#1}{0}}}{\wide@bar{#1}{1}}}
\newcommand*\wide@bar[2]{\if@single{#1}{\wide@bar@{#1}{#2}{1}}{\wide@bar@{#1}{#2}{2}}}
\newcommand*\wide@bar@[3]{%
  \begingroup
  \def\mathaccent##1##2{%
    \let\mathaccent\save@mathaccent
    \if#32 \let\macc@nucleus\first@char \fi
    \setbox\z@\hbox{$\macc@style{\macc@nucleus}_{}$}%
    \setbox\tw@\hbox{$\macc@style{\macc@nucleus}{}_{}$}%
    \dimen@\wd\tw@
    \advance\dimen@-\wd\z@
    \divide\dimen@ 3
    \@tempdima\wd\tw@
    \advance\@tempdima-\scriptspace
    \divide\@tempdima 10
    \advance\dimen@-\@tempdima
    \ifdim\dimen@>\z@ \dimen@0pt\fi
    \rel@kern{0.6}\kern-\dimen@
    \if#31
      \overline{\rel@kern{-0.6}\kern\dimen@\macc@nucleus\rel@kern{0.4}\kern\dimen@}%
      \advance\dimen@0.4\dimexpr\macc@kerna
      \let\final@kern#2%
      \ifdim\dimen@<\z@ \let\final@kern1\fi
      \if\final@kern1 \kern-\dimen@\fi
    \else
      \overline{\rel@kern{-0.6}\kern\dimen@#1}%
    \fi
  }%
  \macc@depth\@ne
  \let\math@bgroup\@empty \let\math@egroup\macc@set@skewchar
  \mathsurround\z@ \frozen@everymath{\mathgroup\macc@group\relax}%
  \macc@set@skewchar\relax
  \let\mathaccentV\macc@nested@a
  \if#31
    \macc@nested@a\relax111{#1}%
  \else
    \def\gobble@till@marker##1\endmarker{}%
    \futurelet\first@char\gobble@till@marker#1\endmarker
    \ifcat\noexpand\first@char A\else
      \def\first@char{}%
    \fi
    \macc@nested@a\relax111{\first@char}%
  \fi
  \endgroup
}
\makeatother

\newcommand{\whj}[1]{}
\newcommand{\SL}[1]{}
\newcommand{\wonmin}[1]{}
\newcommand{\jinwei}[1]{}
\newcommand{\kh}[1]{}
\newcommand{\ye}[1]{}
\newcommand{\YT}[1]{}
\newcommand{\CM}[1]{}
\newcommand{\Sifei}[1]{}
\newcommand{\Yitong}[1]{}
\newcommand{\Hongju}[1]{}
\newcommand{\Collin}[1]{}
\newcommand{\Jinwei}[1]{}
\newcommand{\Pavlo}[1]{}
\newcommand{\Tianfan}[1]{}

\definecolor{Green}{HTML}{E5F8F6}
\definecolor{Blue}{HTML}{DAE3F5}
\definecolor{Purple}{HTML}{ECE3EC}
\definecolor{RPurple}{HTML}{FED8B1}
\definecolor{Yellow}{HTML}{fdffb7}
\definecolor{Grey}{rgb}{0.81176471, 0.81176471, 0.81176471}
\definecolor{DGreen}{rgb}{0.15882353, 0.80980392, 0.14705882}
\definecolor{ngreen}{HTML}{76B900}

\usepackage{tikz}
\usetikzlibrary{arrows.meta,positioning}
\usepackage{pifont}
\usepackage{diagbox}
\usepackage{colortbl}
\usepackage{graphicx}
\usepackage{wrapfig}
\usepackage{mathtools}
\usepackage{placeins}

\usepackage[nameinlink]{cleveref}
\crefname{equation}{Eq.}{Eqs.}
\crefname{figure}{Fig.}{Figs.}
\crefname{section}{Sec.}{Sec.}
\crefname{appendix}{App.}{App.}
\crefname{table}{Tab.}{Tabs.}
\crefname{algorithm}{Algo}{Algo}
\crefname{thm}{Thm}{Thm}
\Crefname{thm}{Thm}{Thm}
\crefname{prop}{Prop}{Prop}

\definecolor{darkred}{rgb}{0.7, 0.0, 0.0}

\title{Scaling Parallel Sequence Models to Foundation-Scale Vision Encoders}

\author{%
\textbf{Yitong Jiang}$^{1,2,\dagger}$\hspace{0.45em}
\textbf{Hongjun Wang}$^{1,3,\dagger}$\hspace{0.45em}
\textbf{Collin McCarthy}$^{1}$\hspace{0.45em}
\textbf{Hanrong Ye}$^{1}$\hspace{0.45em}
\textbf{David Wehr}$^{1}$\hspace{0.45em}
\textbf{Xinhao Li}$^{4}$\hspace{0.45em}
\textbf{Qi Dou}$^{2}$\hspace{0.45em}
\textbf{Tianfan Xue}$^{2}$\hspace{0.45em}
\textbf{Ka Chun Cheung}$^{1}$\hspace{0.45em}
\textbf{Simon See}$^{1}$\hspace{0.45em}
\textbf{Wonmin Byeon}$^{1}$\hspace{0.45em}
\textbf{Ke Chen}$^{1}$\hspace{0.45em}
\textbf{Kai Han}$^{3,\ast}$\hspace{0.45em}
\textbf{Jinwei Gu}$^{1}$\hspace{0.45em}
\textbf{Hongxu Yin}$^{1}$\hspace{0.45em}
\textbf{Pavlo Molchanov}$^{1}$\hspace{0.45em}
\textbf{Jan Kautz}$^{1}$\hspace{0.45em}
\textbf{Sifei Liu}$^{1}$%
\par\vspace{7pt}
{$^1$\,NVIDIA \quad $^2$\,The Chinese University of Hong Kong \quad $^3$\,The University of Hong Kong \quad $^4$\,University of California, San Diego}%
}%

\begin{abstract}
Vision foundation models are bottlenecked by the quadratic cost of self-attention, which caps usable resolution and inflates the price of large-scale pretraining. Subquadratic alternatives such as linear attention and state-space models lower this cost, but they serialize images into 1D token streams and discard the 2D spatial structure that vision relies on. Generalized Spatial Propagation Networks (GSPN) instead propagate context directly on the 2D grid through line-scan recurrences, achieving near-linear complexity and removing positional embeddings---making them a promising primitive for scalable vision. In practice, however, the operator has seen little use as a foundation-scale encoder. We present \textbf{C-GSPN}, a foundation-scale vision encoder built on 2D spatial propagation, realized through three improvements that together make the operator practical: \textbf{(1) a fast GSPN kernel}, \textbf{(2) a more efficient ViT block}, and \textbf{(3) a more efficient training method}.

\emph{(1) A fast GSPN kernel (system efficiency).} The original GSPN has good asymptotic complexity but a hardware-inefficient reference kernel. The fast GSPN kernel fuses thousands of per-step launches into a single warp-specialized CUDA kernel with shared-memory tiling, coalesced access, and a compact multi-channel propagation. It reaches over $90\%$ of peak memory bandwidth, runs up to $40$--$52\times$ faster than the original GSPN implementation, and is already strong as an operator on its own---matching transformer-level ImageNet accuracy and accelerating high-resolution text-to-image synthesis. But a fast kernel alone is not enough: per-layer overhead still dominates at scale, and a brand-new operator does not inherit pretrained attention weights, making from-scratch foundation training very costly.

\emph{(2) A more efficient attention block (architecture efficiency).} C-GSPN performs the propagation in a compressed latent space with fused normalization, cutting per-layer cost while preserving accuracy and turning kernel-level speed into block- and model-level speed.

\emph{(3) A more efficient training method.} Because the new architecture has no inheritable weights, C-GSPN is trained with a two-stage cross-operator distillation recipe---sublayer-wise alignment followed by end-to-end, two-tap feature distillation from an attention teacher---so a foundation-scale model can be obtained cheaply. Distilled with $600$M image--text pairs, C-GSPN matches an isomorphic ViT baseline with $15\%$ fewer parameters, improves ADE20K segmentation by $+2.1\%$, transfers to high resolution with a fraction of the data needed from scratch, and delivers a $4\times$ end-to-end block speedup at 2K with single-pass, tiling-free inference. Together, these three improvements help make 2D spatial propagation a practical basis for foundation-scale vision encoders.
\end{abstract}

\defcitealias{hatamizadeh2024mambavision}{Hatamizadeh et al. (2024)}

\begin{document}
\maketitle

\begingroup
\renewcommand{\thefootnote}{\fnsymbol{footnote}}
\footnotetext[2]{Equal contribution. Work done during an internship at NVIDIA.}
\footnotetext[1]{Corresponding author.}
\endgroup

\vspace{-1.0em}
\noindent
\small
\textbf{Project page:} \href{https://jiangyitong.github.io/cgspn.github.io/}{https://jiangyitong.github.io/cgspn.github.io/}
\par\vspace{1.3em}

\abscontent

\newpage
\tableofcontents
\newpage

\section{Introduction}
\label{sec:intro}

Vision transformers underpin nearly every state-of-the-art vision foundation model: contrastive vision--language encoders such as CLIP~\citep{radford2021learning} and SigLIP~\citep{zhai2023sigmoid}, text-to-image diffusion backbones~\citep{rombach2022high}, and modern detection and segmentation pipelines~\citep{liu2024grounding,kirillov2023segment} all rely on dense, token-wise self-attention to encode visual concepts. The defining weakness of this operator is its quadratic cost in the number of tokens. As resolution grows, token counts surge and attention dominates both memory and latency, so practical systems cap their inputs---SigLIP, for instance, limits images to $512\times512$---and the price of training large models climbs accordingly. Removing this bottleneck without sacrificing the visual quality of attention is a central problem for scaling vision foundation models.

A long line of work makes attention subquadratic through token sparsity~\citep{child2019generatinglongsequencessparse}, local windows~\citep{liu2021swin,dai-etal-2019-transformer}, kernel/low-rank approximations~\citep{katharopoulos2020transformers,choromanski2020rethinking}, IO-aware exact attention~\citep{dao2022flashattention}, and state-space recurrences~\citep{gu2021efficiently,gu2023mamba}. Most of these methods, however, flatten the image into a structure-agnostic 1D sequence, discarding the 2D spatial coherence that is a crucial inductive bias for vision. \emph{Generalized Spatial Propagation Networks} (GSPN)~\citep{Wang2025GSPN} take a different route: they replace 2D self-attention with a learnable line-scan propagation that moves information directly across the rows and columns of the grid. This reduces the effective sequence length to $O(\sqrt{N})$ for $N$ pixels, removes the need for positional embeddings, and, notably, matches or surpasses attention accuracy in prior reports. This makes 2D spatial propagation a \emph{promising primitive} for a scalable, structure-preserving vision encoder.

\textbf{Yet such encoders remain rare.} To our knowledge, spatial propagation has seen little use as a practical foundation-scale vision model. We present \textbf{C-GSPN}, a foundation-scale vision encoder built on 2D spatial propagation, and our central message is that making this work requires improving efficiency at two distinct levels---the propagation \emph{primitive} must be hardware-efficient (a \emph{system} problem), and the \emph{architecture} and its \emph{training} must be efficient enough to actually reach foundation scale---which C-GSPN delivers through three improvements applied in sequence: \textbf{(1) a fast GSPN kernel}, \textbf{(2) a more efficient ViT block}, and \textbf{(3) a more efficient training method}. Kernel speed alone is necessary but not sufficient; the block and the training recipe are what carry it to foundation scale.

\begin{figure}[t]
    \centering
    \includegraphics[width=\linewidth]{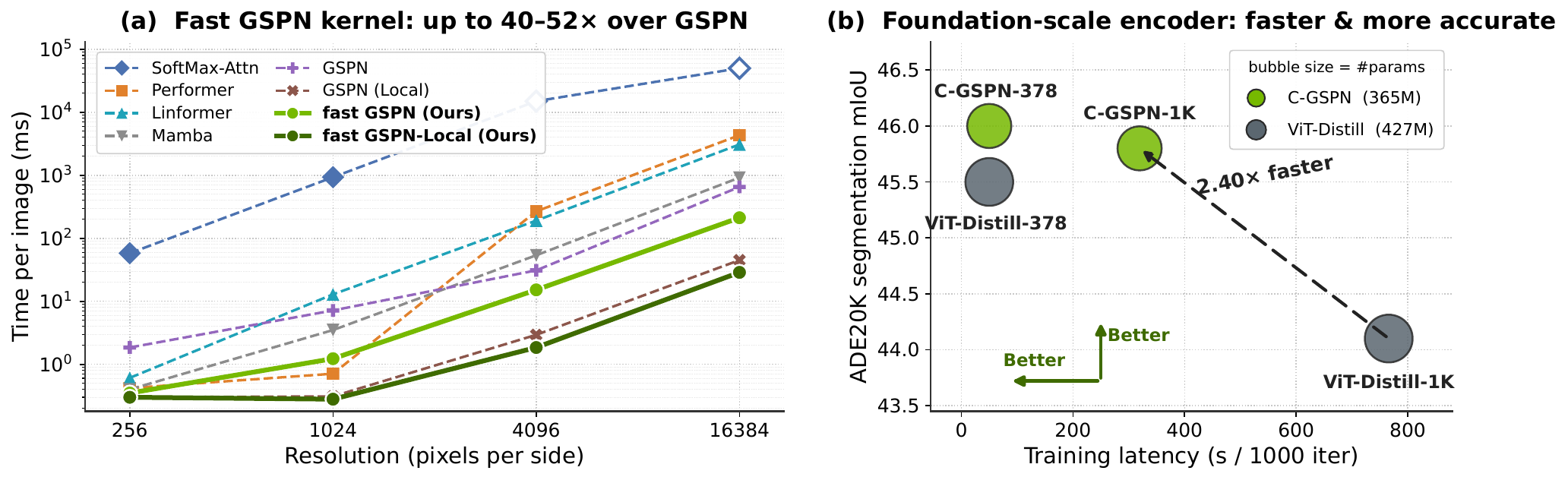}
    \caption{\textbf{One method, C-GSPN, at two levels of efficiency.} \emph{(a)~System efficiency.} The fast GSPN kernel turns the line scan into a single fused, warp-specialized CUDA kernel, running up to $40$--$52\times$ faster than the original GSPN reference kernel across input configurations. \emph{(b)~Architecture \& training efficiency.} Built on this fast kernel, C-GSPN's compressed block and cross-operator distillation scale 2D spatial propagation to a foundation-scale vision encoder, achieving lower training latency and higher ADE20K segmentation accuracy than a distilled ViT at $378$ and 1K resolutions ($2.40\times$ faster at 1K).}
    \label{fig:teaser}
\end{figure}

\textbf{(1) A fast GSPN kernel (Level 1: system efficiency).}
GSPN has good asymptotic complexity but a poorly-mapped reference kernel that launches a tiny CUDA kernel per column step, reaches only $3$--$8\%$ of peak memory bandwidth, repeatedly streams hidden states through global memory, and slows down as channels grow because resident thread blocks saturate GPU concurrency; a modest $256\times256\times1024$ input already exceeds $570$ ms on an A100. C-GSPN's first ingredient is a joint algorithm--system redesign of this kernel---the \emph{fast GSPN kernel}---that (a) fuses all propagation steps into a \emph{single unified CUDA kernel}, (b) introduces a \emph{compact multi-channel propagation} that shares the propagation matrix and projects the scan into a low-dimensional proxy space to bound concurrency pressure, and (c) tunes the grid/block layout with shared-memory caching, coalesced access, 2D channel-parallel blocks, and stream concurrency. The fast GSPN kernel drives the $256\times256\times1024$ case to $11.2$ ms ($52\times$) and a $1024\times1024\times8$ input from $71.4$ ms to $1.8$ ms ($40\times$) at over $90\%$ of peak bandwidth, and is a strong operator in its own right---matching transformer-level ImageNet accuracy and accelerating high-resolution text-to-image synthesis.

\textbf{The bridge --- why a fast kernel is still not enough.}
A fast primitive is necessary but does not yield a foundation-scale encoder. Two obstacles remain. First, even with a near-optimal scan, a propagation \emph{layer} embedded in a ViT block still carries substantial non-propagation overhead (projections, residuals, normalization), which dominates end-to-end latency at the scales foundation encoders use---so kernel speedups do not translate into block-level or model-level speedups for free. Second, and more fundamentally, a foundation encoder must be \emph{trained}: attention-based encoders are pretrained on tens of billions of pairs (e.g., SigLIP-v2~\citep{tschannen2025siglip} uses $40$B), and because spatial propagation is a \emph{new operator with no inheritable attention weights}, training one from scratch at that scale is very costly. Efficiency therefore has to be solved again at the architecture and training levels.

\textbf{(2) A more efficient ViT block and (3) a more efficient training method (Level 2: architecture and training efficiency).}
C-GSPN resolves both obstacles. \emph{(2) The block.} Guided by the kernel's concurrency analysis---propagation latency is governed by the channel dimension---C-GSPN performs the propagation in a \emph{compressed latent space}: it down-projects to a small set of latent channels, runs the four-directional fast scan there with a fused row-stochastic normalization kernel, and up-projects once, while stripping the inherited attention-era overhead. This cuts layer-level propagation latency by nearly $10\times$ at high resolution with no loss of accuracy and yields a $13.7\times$ speedup of the full GSPN layer at 1K. \emph{(3) The training.} Because the block does not inherit attention weights, we make foundation-scale training cheap with a \emph{two-stage cross-operator distillation} recipe: a sublayer-wise stage aligns each propagation sublayer with its teacher's attention sublayer for a strong initialization, then end-to-end distillation with two supervision taps per block (post-propagation and post-block) and lightweight feature adaptors bridges the operator gap. Distilled from an attention teacher on $600$M image--text pairs, C-GSPN matches an isomorphic ViT$\to$ViT baseline with $15\%$ fewer parameters ($63.3$ vs.\ $63.5$ macro average), improves ADE20K segmentation by $+2.1\%$, and---because propagation needs no positional embeddings---transfers to higher resolutions through a cheap upsampling self-distillation curriculum using a fraction of the data required to train from scratch. End to end, C-GSPN reduces ViT block latency by $2\times$ at 1K and $4\times$ at 2K with single-pass, tiling-free inference.

\noindent\textbf{Contributions.}
\begin{itemize}
    \item \textbf{C-GSPN, a foundation-scale vision encoder built on 2D spatial propagation.} To our knowledge, one of the first studies to scale a subquadratic, structure-preserving spatial operator to CLIP/SigLIP-style pretraining while preserving zero-shot ability and improving dense prediction---realized through three efficiency improvements below.
    \item \textbf{(1) A fast GSPN kernel.} A joint algorithm--system redesign of the propagation kernel (single fused kernel, compact multi-channel propagation, warp/shared-memory-aware execution) that reaches $>$$90\%$ of peak bandwidth and is $40$--$52\times$ faster than the original GSPN reference kernel while preserving accuracy.
    \item \textbf{(2) A more efficient ViT block.} A compressed latent-space propagation block with fused normalization that removes inherited attention-era overhead and delivers large high-resolution block-level speedups.
    \item \textbf{(3) A more efficient training method.} A progressive, two-stage cross-operator distillation recipe (sublayer-wise alignment then dual-tap end-to-end distillation with feature adaptors) that makes foundation-scale, positional-embedding-free training and tiling-free high-resolution transfer affordable for an operator with no inheritable weights.
    \item \textbf{Comprehensive evaluation} spanning kernel profiling, ImageNet classification, text-to-image synthesis, and CLIP-scale zero-shot, segmentation, and detection benchmarks, evaluating C-GSPN at three granularities against prior baselines (original GSPN, softmax/FlashAttention, ViT teachers).
\end{itemize}

\section{Related Work}
\label{sec:related_work}

\paragraph{Efficient and subquadratic attention.}
Transformers~\citep{vaswani2017attention} are foundational to modern vision and language models, but their $O(N^2)$ cost in sequence length is a persistent efficiency barrier. IO-aware exact attention such as FlashAttention~\citep{dao2022flashattention,dao2023flashattention2,shah2024flashattention} fuses the attention pipeline and optimizes memory traffic to raise throughput, but its latency still scales quadratically with tokens at high resolution. \emph{Sparsity- and window-based} designs (Longformer, BigBird, Swin) restrict attention to local windows plus a few global tokens for near-linear scaling, at the cost of sensitivity to the chosen sparsity pattern~\citep{Beltagy2020Longformer,zaheer2020big,liu2021swin}. \emph{Kernelized and low-rank} approaches---Linear Transformers, Performer, Nystr\"omformer, Linformer---approximate the softmax to gain linear complexity, with accuracy that depends on the feature map, rank, or landmark scheme~\citep{katharopoulos2020transformers,choromanski2020rethinking,xiong2021nystromformer,wang2020linformer}. These methods reduce the asymptotic cost but, like most efficient-attention variants, treat the image as a 1D token stream.

\paragraph{State-space and 1D/2D sequence models.}
Recurrent models---LSTMs~\citep{hochreiter1997long}, GRUs~\citep{chung2014empirical}, and 2D-LSTMs~\citep{graves2007multi,byeon2015scene}---process data through non-linear transitions but are limited by sequential execution and long-range gradient instability~\citep{hochreiter1991untersuchungen,pascanu2013difficulty}. State-space models (SSMs) such as S4~\citep{gu2021efficiently} and Mamba~\citep{gu2023mamba} offer linear-time sequence operators with selective scanning, and several works adapt them to vision by linearizing 2D images into 1D scans~\citep{nguyen2022s4nd,baron20232,zhu2024ViM,liu2024vmamba,li2024mamba}. While efficient, this serialization compromises the spatial relationships that vision tasks depend on, and adapting 1D recurrences to high-resolution vision typically requires extra 2D inductive bias or hierarchical design.

\paragraph{Spatial propagation networks.}
The Spatial Propagation Network (SPN)~\citep{liu2017learning} pioneered learnable linear propagation on 2D data, initially as a refinement layer on top of CNNs for sparse-to-dense prediction. SPN's sequential, single-direction processing limits efficiency and long-range reach. GSPN~\citep{Wang2025GSPN} advances this idea with parallel row/column-wise propagation in four directions, learning input-dependent affinities while maintaining stability and dense pairwise connectivity at $O(\sqrt{N})$ effective depth, and positions spatial propagation as a competitive alternative to ViT and Mamba backbones. Our work builds directly on GSPN: C-GSPN first makes its propagation kernel hardware-efficient, then scales the operator to foundation-model pretraining through an efficient ViT block and a cross-operator distillation recipe.

\paragraph{Foundation-model and cross-operator distillation.}
Training vision foundation models from scratch is increasingly infeasible---recent vision--language encoders require tens of billions of image--text pairs and massive compute. A practical alternative transfers knowledge from strong pretrained attention encoders into more efficient students. Knowledge distillation is the standard mechanism: early work studied feature and attention transfer for CNNs~\citep{zagoruyko2016paying}, DeiT~\citep{touvron2021training} demonstrated ViT-to-ViT distillation at scale, and later studies provided guidance on intermediate supervision and layer alignment for stable transformer KD~\citep{yang2022vitkd}. Beyond isomorphic pairs, distillation has been used across operator families---compressing attention-heavy models into subquadratic approximations~\citep{bick2024transformers} or distilling SSMs into transformer backbones~\citep{li2025matvlm}---though usually at modest scale. We use distillation not for size compression but for \emph{cross-operator} transfer from attention teachers to spatial-propagation students, via staged sublayer-wise initialization followed by end-to-end training, enabling foundation-scale, high-resolution encoders with a subquadratic core.

\section{Background: 2D Spatial Propagation}
\label{sec:background}

We first establish the propagation operator and a single notation used throughout the paper. All parts of our method---the fast GSPN kernel (\cref{sec:gspn2}) and the compressed ViT block and training recipe (\cref{sec:cgspn})---operate on this same primitive, so we introduce it once here. Throughout, an input feature map is $x \in \mathbb{R}^{B \times C \times H \times W}$ with batch size $B$, channel count $C$, and spatial size $H \times W$; we write $N = HW$ for the number of pixels/tokens. For naming, we use \emph{GSPN} for the original method of \citet{Wang2025GSPN} and its reference CUDA kernel (the prior-work baseline), \emph{fast GSPN} for our fast fused kernel (the first improvement, \cref{sec:gspn2}), and \emph{C-GSPN} for our overall foundation-scale encoder---whose fast GSPN kernel, compressed block, and distillation recipe are the three improvements developed in this paper.

\paragraph{Line-scan recurrence.}
GSPN performs 2D spatial modeling through row-by-row (or column-by-column) linear propagation: one spatial dimension is processed sequentially while all positions within each step are updated in parallel. Taking the top-to-bottom pass as an example, let $i \in [0, H{-}1]$ index rows and $c$ index channels, with hidden state $h_{i,:,c} \in \mathbb{R}^{W}$, input row $x_{i,:,c} \in \mathbb{R}^{W}$, an input-dependent scaling vector $\lambda_{i,:,c} \in \mathbb{R}^{W}$, and a propagation matrix $w_{i,c} \in \mathbb{R}^{W \times W}$. The per-row, per-channel recurrence is
\begin{equation}
h_{i,:,c} = w_{i,c}\, h_{i-1,:,c} + \operatorname{Diag}(\lambda_{i,:,c})\, x_{i,:,c},
\label{eq:gspn_recurrence}
\end{equation}
with $h_{0,:,c}$ initialized from $x_{0,:,c}$, and an output gating
\begin{equation}
y_{i,:,c} = u_{i,:,c} \odot h_{i,:,c},
\label{eq:gspn_output}
\end{equation}
where $u_{i,:,c} \in \mathbb{R}^{W}$ is a learnable vector. All parameters $\lambda$, $w$, and $u$ are input-dependent.

\paragraph{Stability and connectivity.}
To satisfy the Stability--Context Condition of \citet{Wang2025GSPN}, each $w_{i,c}$ is row-stochastic (its rows sum to one), which promotes numerical stability while still allowing long-range context. We parameterize this by normalizing the nonzero connections over the neighbor set $\mathcal{N}(j)$ of position $j$ in row $i$:
\begin{equation}
w_{i,c}(j,k) = \frac{\sigma(\tilde w_{i,c}(j,k))}{\sum_{k' \in \mathcal{N}(j)} \sigma(\tilde w_{i,c}(j,k'))}.
\label{eq:row_stochastic_norm}
\end{equation}
With the common tridiagonal neighborhood $\mathcal{N}(j) = \{j{-}1, j, j{+}1\}$, each row of $w_{i,c}$ has three nonzero entries, so every pixel connects to only three neighbors in the previous row (e.g., top-left, top-center, top-right). A single pass therefore connects pixels within a local region; running four complementary directional passes---top-to-bottom, bottom-to-top, left-to-right, right-to-left---yields dense pairwise connectivity across the image while learning only three coefficients per pixel per pass.

\paragraph{Complexity and the attention analogy.}
A row pass requires $O(H)$ sequential steps with all $W$ positions computed in parallel (symmetrically $O(W)$ for a column pass), giving an effective sequential depth of $O(\max(H,W)) = O(\sqrt{N})$ for a square map. Stacking the per-channel recurrence over all rows reveals the operator's relationship to attention. Concatenating hidden states and inputs into vectors $H_v, X_v$ and writing $\Lambda_i$ for the (block-diagonal) input scaling, \cref{eq:gspn_recurrence} expands to a block lower-triangular form
\begin{small}
\begin{equation}
H_v =
\left[
\begin{matrix}
\Lambda_1 & 0 & \cdots & \cdots & 0 \\
w_2\Lambda_1 & \Lambda_2 & 0 & \cdots & 0 \\
w_3 w_2 \Lambda_1 & w_3 \Lambda_2 & \Lambda_3 & 0 & \cdots \\
\vdots & \vdots & \vdots & \ddots & \vdots \\
(\prod_{k=2}^{L} w_k)\Lambda_1 & (\prod_{k=3}^{L} w_k)\Lambda_2 & \cdots & w_L\Lambda_{L-1} & \Lambda_L \\
\end{matrix}
\right] X_v = G\, X_v,
\label{eq:global}
\end{equation}
\end{small}
where each block $G_{ij}$ specifies how input slice $x_j$ contributes to output $h_i$---directly analogous to an attention affinity matrix. The products $\prod_{\tau=j+1}^{i} w_\tau$ play the role of normalized affinities, while the $\Lambda_j$ inject per-position value gating. This view, which the fast GSPN kernel makes explicit through channel-shared weights (\cref{sec:gspn2}), motivates treating spatial propagation as a drop-in, attention-like mixing operator.

\paragraph{GPU execution model.}
Both contributions are co-designed with GPU hardware, so we briefly recall the relevant characteristics of modern NVIDIA GPUs (e.g., the A100). A CUDA kernel is launched as a grid of thread blocks; each block holds up to $1024$ threads grouped into $32$-thread warps, the basic scheduling unit on the $108$ Streaming Multiprocessors (SMs). Throughput is maximized when occupancy---the fraction of active warps per SM---is high, subject to per-SM register and shared-memory limits. Crucially, each SM can host only a bounded number of resident blocks (up to $32$ on the A100), so roughly $108 \times 32 \approx 3{,}500$ blocks run concurrently; once the workload exceeds this, additional blocks queue and latency grows. As we show next, the line scan's mapping onto this model---one block per $(B, C, \text{chunk})$ slice---is exactly what makes the naive GSPN kernel slow and what fast GSPN is designed to fix. A fuller GPU and kernel-execution primer is given in \cref{app:gpu}.

\section{Level 1: A Fast GSPN Kernel (Level 1: System Efficiency)}
\label{sec:gspn2}

The first ingredient of C-GSPN makes the line-scan primitive of \cref{sec:background} fast on real hardware. This is the enabling foundation layer for the rest of the encoder (\cref{sec:cgspn}): without a hardware-efficient scan, architectural design alone is unlikely to make spatial propagation practical at scale. We begin from the original GSPN reference kernel to expose its bottlenecks, then present a joint algorithm--system redesign---\emph{the fast GSPN kernel}---organized around three principles: (1) a single-kernel propagation scheme that eliminates redundant launches, (2) a compact channel propagation that shares weights and compresses the channel axis to relieve concurrency pressure, and (3) a CUDA execution layout that exploits shared memory, coalesced access, and stream-level parallelism. As we show in \cref{sec:experiment}, the fast GSPN kernel is already a strong operator on its own; \cref{subsec:gspn2_bridge} then explains why these kernel-level wins, though necessary, are still not sufficient for a foundation-scale encoder---motivating the block and training improvements that follow.

\begin{figure}[t]
    \centering
    \includegraphics[width=\textwidth, trim=0 36 0 0, clip]{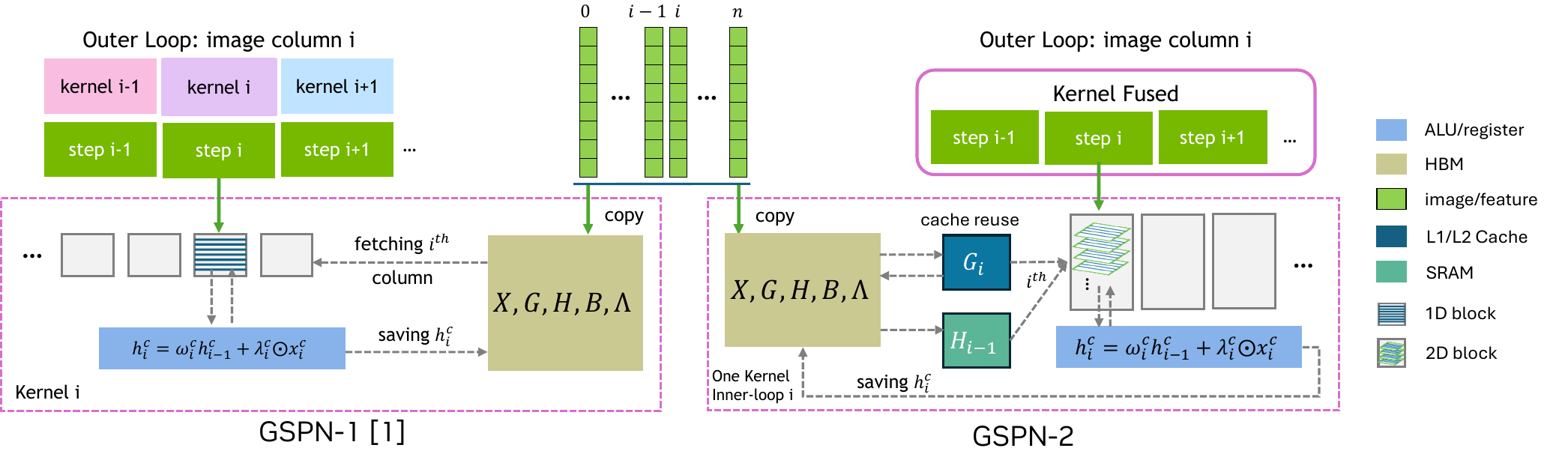}\\[2pt]
    \makebox[\textwidth][l]{%
      \hspace*{\dimexpr 0.211\textwidth-1.6cm\relax}%
      \makebox[3.2cm][c]{\normalsize\textbf{(a) GSPN kernel}}%
      \hspace*{\dimexpr 0.441\textwidth-3.2cm\relax}%
      \makebox[3.2cm][c]{\normalsize\textbf{(b) fast GSPN kernel}}%
    }
    \caption{\textbf{From the GSPN kernel to the fast GSPN kernel.} \textbf{(a)~GSPN kernel:} the original reference kernel launches a separate, lightweight kernel per image column, computing $h_i^c = \omega_i^c h_{i-1}^c + \lambda_i^c \odot x_i^c$ and shuttling intermediate states $X,G,H,B,\Lambda$ through global memory (HBM) every step, with no on-chip reuse---the bottleneck. \textbf{(b)~fast GSPN kernel:} a single fused kernel runs the outer scan with an inner loop over columns, caching and reusing $h_{i-1}^c$, $G_i$, and other temporaries on chip (cache/registers) to minimize HBM traffic.}
    \label{fig:cuda_kernel}
\end{figure}

\paragraph{The GSPN baseline and its bottlenecks.}
GSPN maps the recurrence of \cref{eq:gspn_recurrence} to CUDA by iterating sequentially over the propagation dimension (e.g., height $H$) while parallelizing across the orthogonal dimension (width $W$). To respect the sequential dependency along rows $i = 0,\dots,H{-}1$, it launches a separate, lightweight kernel for each step or small chunk (\cref{fig:cuda_kernel}a). Within each kernel, computation is parallelized across width $W$, batch $B$, and channels $C$ by flattening them into a 1D grid of thread blocks (typically \texttt{blockDim.x = 512}). This design has three compounding inefficiencies: (i) the kernel-launch overhead from thousands of separate launches prevents the SMs from staying busy; (ii) every step reloads inputs $x$, previous hidden states $h_{i-1,:,c}$, weights $w_{i,c},\lambda_{i,:,c}$, and writes outputs through global memory (HBM) with no on-chip reuse and poor coalescing; and (iii) the flat 1D mapping ignores warp-level scheduling, so locality and occupancy are poor and runtime grows with the channel count. Profiling confirms GSPN reaches only $3$--$8\%$ of peak memory bandwidth, and a $256\times256\times1024$ input exceeds $570$ ms on an A100---eroding much of the subquadratic advantage and motivating a single-kernel, shared-memory redesign.

\subsection{A Single-Kernel Design}
\label{subsec:single_kernel}

\begin{figure}[t]
  \centering
  \includegraphics[width=0.58\textwidth]{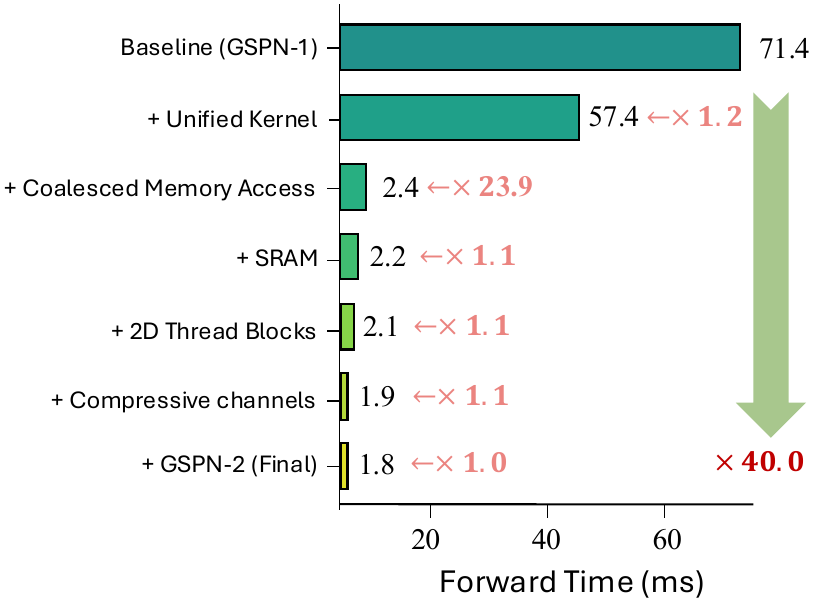}
  \caption{\textbf{Step-by-step optimization of the GSPN CUDA kernel.} Each bar shows the cumulative reduction in forward time (ms) from the original GSPN baseline. The final fast GSPN kernel achieves a $40.0\times$ speedup.}
  \label{fig:journey}
\end{figure}

\paragraph{Kernel fusion.}
We consolidate the thousands of per-step launches into a \emph{single, unified CUDA kernel} that processes the entire outer scan (e.g., all columns in a left-to-right pass) \emph{inside} the kernel, while still parallelizing across batch, channels, and the orthogonal spatial axis (\cref{fig:cuda_kernel}b). Eliminating the micro-launches alone yields an immediate, if modest, gain---fusing the multi-kernel pipeline into one kernel gives roughly a $1.2\times$ speedup even before any memory or algorithmic optimization (\cref{fig:journey})---and, more importantly, exposes the inner loop where hidden states can be reused on chip.

\paragraph{Block and grid configuration.}
GSPN's flat 1D grid (\texttt{blockDim.x = 512}) spreads threads linearly across combinations of batch $B$, channels $C$, height $H$, and chunk index $k_{\text{chunk}}$, yielding poor locality and warp utilization. The fast GSPN kernel instead indexes the grid by the tuple $(\text{chunk}, b, c)$, so each block owns one $(\text{chunk}, b, c)$ slice and processes a full spatial column along height. The grid contains $k_{\text{chunk}} \times B \times C$ blocks, realized as a 1D or 3D grid to respect CUDA's per-axis limits, and is launched once. Each block uses up to $1024$ threads along height: for $H \le 1024$ one thread handles one row at full occupancy, and for $H > 1024$ threads stride over multiple rows. This mapping removes per-thread index-unpacking and improves cache locality and occupancy.

\subsection{Compact Channel Propagation}
\label{sec:gspnv2_theory}

\paragraph{Concurrency saturation.}
A key bottleneck in GSPN is GPU concurrency saturation: the number of active blocks is proportional to $k_{\text{chunk}} \times B \times C$, and once it exceeds the hardware's resident-block capacity (about $3$--$4$K on an A100, \cref{sec:background}), kernel time grows linearly as blocks queue. This is exactly why GSPN loses its near-constant scaling on high-dimensional feature maps with thousands of channels.

\paragraph{Channel-shared weights and the attention view.}
To address this, the fast GSPN kernel introduces a \emph{compact multi-channel propagation} that reduces effective channel concurrency while preserving expressive multi-channel behavior. Instead of a per-channel matrix $w_{i,c}$ as in GSPN, it uses a single matrix $w_i \in \mathbb{R}^{W\times W}$ shared across all channels for each column $i$:
\begin{equation}
h_{i,:,c} = w_i\, h_{i-1,:,c} + \operatorname{Diag}(\lambda_{i,:,c})\, x_{i,:,c}.
\label{eq:2dsp}
\end{equation}
Here $w_i$ governs spatial propagation along the column while $\lambda_{i,:,c}$ preserves per-channel modulation. This sharply reduces the parameters for propagation while keeping the same functional structure: stacking all channels, the full recurrence $h_i = W_i h_{i-1} + \Lambda_i x_i$ still holds, now with channel-shared $w_i$. Through the block lower-triangular expansion of \cref{eq:global}, the shared $w_i$ play the role of an attention-style affinity matrix over positions, while the channel-specific $\Lambda_j$ act as value gating---so in the single-channel case the kernel is precisely an attention-like process with learnable spatial affinities, aligning the operator more closely with the attention it replaces.

\paragraph{Compressive proxy dimension.}
To further relieve saturation when $B \times C$ is large, the kernel compresses the channel axis before propagation. We project $x \in \mathbb{R}^{B\times C\times H\times W}$ to a proxy $x_{\mathrm{proxy}} \in \mathbb{R}^{B\times C_{\mathrm{proxy}}\times H\times W}$ with $C_{\mathrm{proxy}} \ll C$ (e.g., $C_{\mathrm{proxy}}{=}8$), apply the same column-wise recurrence in the proxy space with the shared $w_i$, and expand back to $C$ with a learned $1\times1$ projection. This reduces the grid from $k_{\text{chunk}}\times B\times C$ to $k_{\text{chunk}}\times B\times C_{\mathrm{proxy}}$, shrinking the number of simultaneously scheduled slices and keeping the active-block count within the hardware concurrency regime. We choose $C_{\mathrm{proxy}}$ to minimize the active-block budget and delay entry into the post-saturation, near-linear regime; even when very large $B$ makes the plateau unavoidable, the compression still reduces queueing and improves SM utilization while preserving multi-channel expressiveness. We show in \cref{app:lowrank} that this proxy is effectively a low-rank factorization with minimal accuracy cost.

\subsection{Efficient CUDA Scaling under Large Block-Slice Loads}
\label{subsec:gspn2_cuda_opts}
The single-kernel design and compact propagation are complemented by CUDA-level optimizations---grid/block reconfiguration and on-chip memory strategies---that keep execution efficient even when the block count $k_{\text{chunk}}\times B\times C$ is large.

\paragraph{Shared memory for hidden states.}
We cache the previous step's hidden state $h_{i-1}$ in on-chip shared memory to cut redundant HBM reads. Threads in a block cooperatively process a tile---a small subset of spatial positions or channel slices---and reuse the cached hidden state, lowering latency when accesses overlap along a column. The gain depends on configuration: it is largest when per-tile reuse is high and the shared-memory footprint fits per-block limits with few bank conflicts, and it diminishes when L1/L2 already covers the working set. We therefore enable shared-memory caching selectively and tune tile size and \texttt{cSlice} to balance reuse against occupancy.

\paragraph{2D blocks for channel-parallel propagation.}
Building on the 1D block of \cref{subsec:single_kernel}, we add a second block dimension \texttt{cSlice}, so $\texttt{blockDim} = (H, \texttt{cSlice})$. Within a block, \texttt{threadIdx.x} indexes spatial positions along a column (up to $H$) while \texttt{threadIdx.y} spans a small group of channel slices, letting one block process several channels of the same column in parallel. Aligning computation and memory access across both spatial and channel dimensions improves occupancy and reduces latency relative to the 1D layout.

\paragraph{Coalesced memory access.}
We arrange $x_i$, $h_i$, and $w_i$ contiguously so that consecutive threads in a warp touch adjacent addresses when reading or writing. The hardware then coalesces per-thread transactions into wide memory operations, fully using bandwidth and eliminating the scattered, high-latency accesses of GSPN. This coalesced layout contributes the single largest CUDA-level speedup (\cref{fig:journey}).

\paragraph{Stream-based concurrency.}
For multi-directional propagation, the fast GSPN kernel runs each of the four directional passes on a separate, non-blocking CUDA stream, overlapping their execution to keep more SMs active---most effective when the passes have similar compute and memory footprints. When a grid dimension exceeds CUDA's per-axis limit of $65{,}535$, it automatically issues multiple launches with offset indexing without interrupting stream concurrency.

\subsection{Why the Fast Kernel Is Not Enough}
\label{subsec:gspn2_bridge}
The fast GSPN kernel makes the propagation primitive fast---up to $52\times$ over the original GSPN kernel at over $90\%$ of peak bandwidth (\cref{sec:experiment})---and competitive on classification and generation. But a fast primitive does not, by itself, yield a foundation-scale vision encoder, for two reasons that the rest of the paper resolves. \emph{(i) Layer- and model-level overhead.} Once the scan is this cheap, the surrounding non-propagation components of a ViT block---projections, residuals, and especially the sigmoid-based row-stochastic normalization---dominate end-to-end latency at the resolutions and channel widths foundation encoders use; a fast kernel alone does not translate into a fast \emph{block}. \emph{(ii) Trainability of a new operator.} A foundation encoder must be pretrained at large scale, yet spatial propagation is a \emph{new operator with no inheritable attention weights}, so training one from scratch at CLIP/SigLIP scale is very costly. Efficiency must therefore be addressed again at the architecture and training levels. This is exactly the gap C-GSPN closes in \cref{sec:cgspn}.

\section{Level 2: Architecture and Training Efficiency for Foundation Scale (C-GSPN)}
\label{sec:cgspn}

\begin{figure*}[t]
    \centering
    \begin{minipage}[t]{0.20\textwidth}
        \centering
        \includegraphics[width=\linewidth]{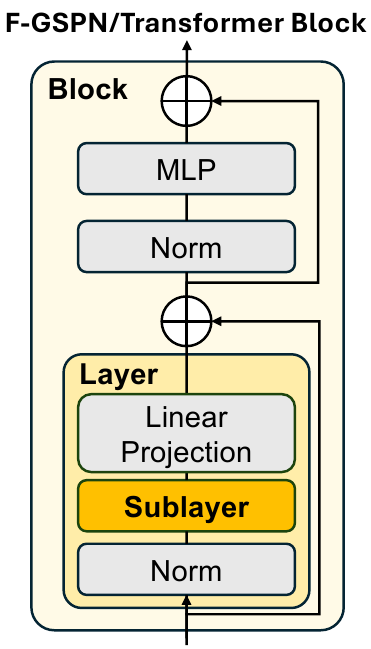}\\[2pt]
        {\footnotesize (a) Block / Layer / Sublayer}
    \end{minipage}\hfill
    \begin{minipage}[t]{0.78\textwidth}
        \centering
        \includegraphics[width=\linewidth]{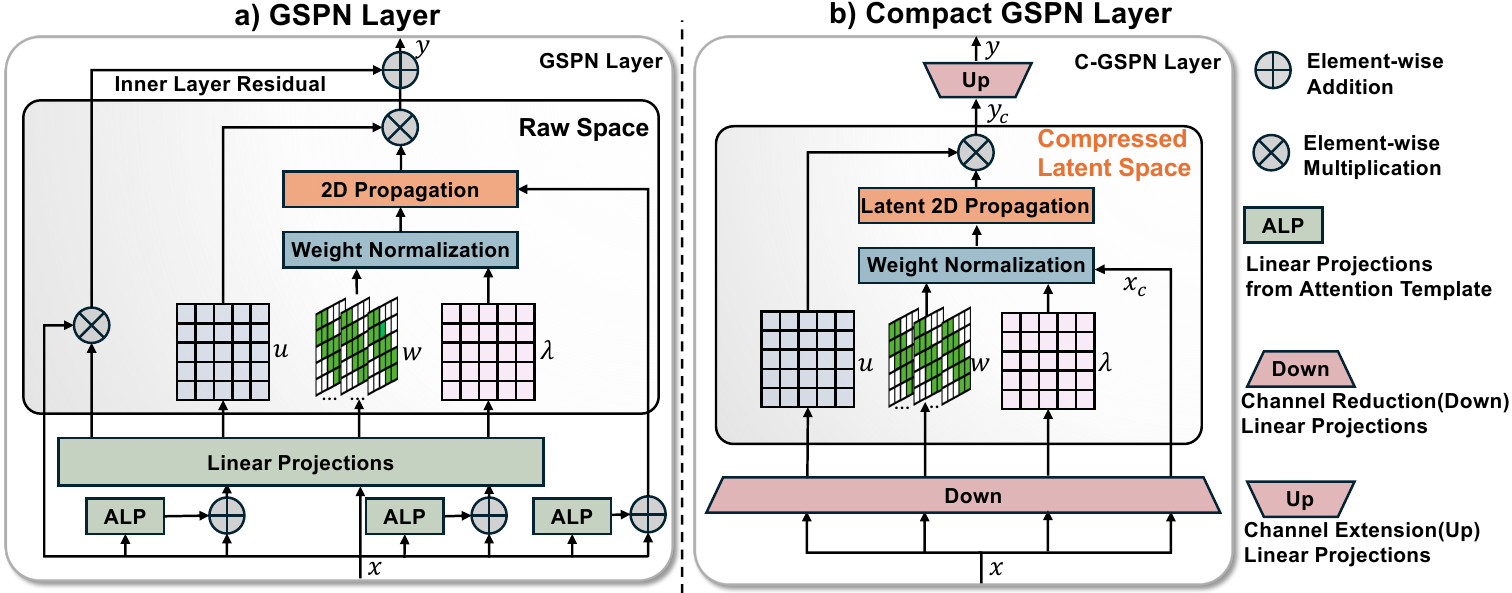}\\[2pt]
        {\footnotesize (b) GSPN layer (left) vs.\ C-GSPN layer (right)}
    \end{minipage}
    \caption{\textbf{C-GSPN architecture overview.} \emph{(a)} C-GSPN follows the ViT hierarchy of \textit{block} $\supset$ \textit{layer} $\supset$ \textit{sublayer}, replacing only the attention layer. \emph{(b)} The original GSPN layer operates in raw channel space and keeps the extra projections and residuals inherited from the attention template (Improvement~2's target); C-GSPN propagates in a compressed latent space with fused row-stochastic normalization and removes the redundant projections/residuals, yielding a lighter, faster layer. For clarity, the final propagation pass at the end of the layer is omitted.}
    \label{fig:structure}
\end{figure*}

Level 1 gave us a fast scan but, as \cref{subsec:gspn2_bridge} argued, not a foundation encoder: layer-level overhead still dominates at scale, and a new operator does not inherit attention weights, making cheap training difficult. \textbf{C-GSPN} is our second level of efficiency, and the headline contribution of the paper: a foundation-scale vision encoder that resolves both obstacles through architecture and training design. It has two pillars. \textbf{(a) A more efficient ViT block} (\cref{sec:efficiency}) runs the fast GSPN kernel of \cref{sec:gspn2} inside a compressed latent space with fused normalization, so kernel-level speed finally translates into block- and model-level speed. \textbf{(b) Fast training of a new architecture} (\cref{sec:distill}) uses cross-operator distillation to obtain a foundation-scale model without the from-scratch cost that a non-inheritable operator would otherwise incur, with high-resolution transfer following the same recipe (\cref{sec:highres-transfer}). We align C-GSPN with the ViT structure and adopt consistent terminology: a \textit{sublayer} is the latent-space 2D propagation unit, analogous to the scaled dot-product attention sublayer; a \textit{layer} is a full C-GSPN layer (\cref{fig:structure}), paralleling a multi-head attention layer; and a \textit{block} is a transformer block where the attention layer is replaced by a C-GSPN layer (\cref{fig:structure}).

\subsection{A More Efficient ViT Block: Latent-Space Propagation}
\label{sec:efficiency}

\begin{figure}[t]
    \centering
    \includegraphics[width=0.6\textwidth]{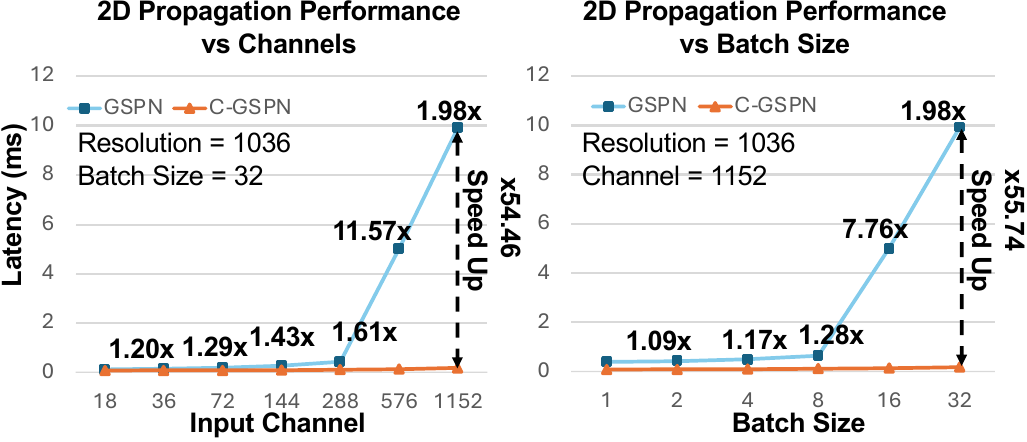}
    \caption{\textbf{Propagation sublayer latency, original GSPN vs.\ C-GSPN}, under increasing channels (left) and batch size (right) at 1K resolution. Original GSPN spikes as $C$/$B$ grow due to GPU concurrency limits; C-GSPN's latent-space propagation remains flat, yielding large speedups.}
    \label{fig:compression-b}
\end{figure}

The compact-channel principle of the fast GSPN kernel (\cref{sec:gspnv2_theory}) suggests where the encoder's cost lies. A raw GSPN sublayer propagates features independently across each of the $C$ channels of $x \in \mathbb{R}^{B\times C\times H\times W}$, running the four directional scans of \cref{eq:gspn_recurrence}. Because each SM hosts only a bounded number of resident blocks, once $B$ or $C$ grows the excess slices serialize and latency spikes despite the theoretical parallelism. \cref{fig:compression-b} illustrates this: propagation latency is flat at small $B/C$ but jumps sharply once concurrency saturates (e.g., a $11.57\times$ increase when $C$ grows from $288$ to $576$, and $7.76\times$ when $B$ grows from $8$ to $16$).

\paragraph{Latent-space 2D propagation.}
To stay below the concurrency wall, C-GSPN moves propagation into a compressed latent space---the learned, end-to-end-trained analog of the kernel's proxy compression. Let $s>1$ be a compression factor and $C_c = \lfloor C/s \rfloor$ the latent channel count. We introduce per-location ($1\times1$ convolution) projections $P_{\downarrow}:\mathbb{R}^{C}\!\to\!\mathbb{R}^{C_c}$ and $P_{\uparrow}:\mathbb{R}^{C_c}\!\to\!\mathbb{R}^{C}$:
\begin{equation}\label{eq:latent_down}
\mathbf{x}_c = P_{\downarrow}(\mathbf{x}) \in \mathbb{R}^{B\times C_c\times H\times W}.
\end{equation}
Propagation parameters are generated directly in the latent channel space,
\begin{equation}\label{eq:latent_params}
u = L_u(\mathbf{x}_c),\quad \lambda = L_\lambda(\mathbf{x}_c),\quad \tilde w = L_w(\mathbf{x}_c),
\end{equation}
where $u,\lambda \in \mathbb{R}^{B\times C_c\times H\times W}$ and $\tilde w \in \mathbb{R}^{B\times C_c\times H\times W\times 3}$, and $L_u, L_\lambda, L_w$ are $1\times1$ convolutions predicting per-position parameters. For a top-to-bottom pass and latent channel $\tilde c$, the recurrence mirrors \cref{eq:gspn_recurrence} but operates entirely on the latent channels with the row-stochastic normalization of \cref{eq:row_stochastic_norm}:
\begin{equation}
\label{eq:latent_recurrence}
h_{i,:,\tilde c} = w_{i,\tilde c}\, h_{i-1,:,\tilde c} + \operatorname{Diag}(\lambda_{i,:,\tilde c})\, x_{c,\,i,:,\tilde c}, \qquad
y_{i,:,\tilde c} = u_{i,:,\tilde c} \odot h_{i,:,\tilde c}.
\end{equation}
We run all four directional scans in latent space and up-project only once at the end:
\begin{equation}
\label{eq:latent_up}
\mathbf{y}_c = \text{Prop}_{2\mathrm{D}}(\mathbf{x}_c; u,\lambda,w), \qquad
\mathbf{y} = P_{\uparrow}(\mathbf{y}_c) \in \mathbb{R}^{B\times C\times H\times W}.
\end{equation}
This reduces the effective grid from $B\!\times\!C$ to $B\!\times\!C_c$, lowering per-SM pressure and avoiding serialization. At 1K resolution, latency stays nearly flat across channels and batch sizes; compared to the raw-space kernel, latent-space propagation attains $54.46\times$ speedup at $C{=}1152$ and $55.74\times$ at $B{=}32$ (\cref{fig:compression-b}). Because $\tilde w$ lives in the latent channels, the normalization of \cref{eq:row_stochastic_norm} is evaluated over $C_c$ rather than $C$, giving an additional $38.9\times$ speedup in weight normalization.

\begin{figure*}[t]
    \centering
    \begin{minipage}[t]{0.47\textwidth}
        \centering
        \includegraphics[width=\linewidth]{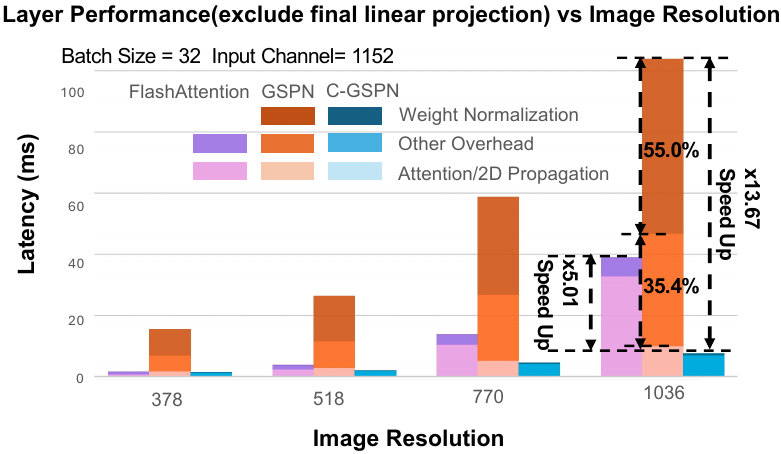}
    \end{minipage}\hfill
    \begin{minipage}[t]{0.45\textwidth}
        \centering
        \includegraphics[width=\linewidth]{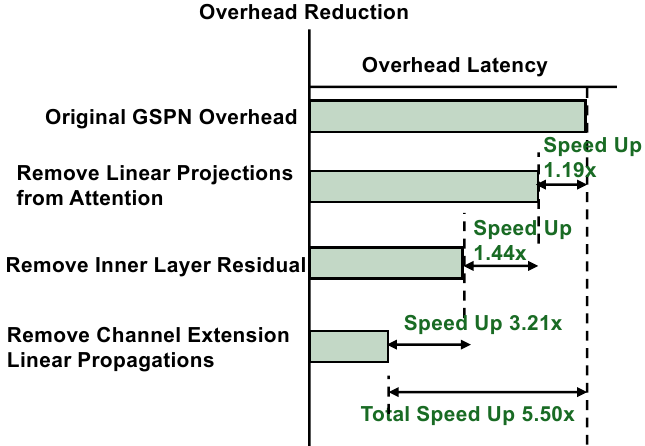}
    \end{minipage}
   \caption{\textbf{Left}: Block latency vs.\ image resolution ($B{=}32$, $C{=}1152$); the original GSPN is dominated by weight normalization and other overhead at high resolution, which C-GSPN substantially reduces. \textbf{Right}: Overhead reduction at 1K from removing (1) additional linear projections, (2) the inner-layer residual, and (3) channel-extension projections; cumulative speedup $\approx\!5.5\times$.}
    \label{fig:module_performance}
    \label{fig:overhead}
\end{figure*}

\paragraph{Non-propagation overhead reduction.}
Is the propagation sublayer really the bottleneck? For softmax attention at high resolution, yes; but GSPN's propagation is already efficient, so at low and medium resolutions the runtime is dominated by the \emph{non-propagation} components. At 1K, these parts cost $9.6\times$ more latency than the core propagation (\cref{fig:module_performance}). We therefore strip overhead inherited from the attention-era template by removing (i) the inner-layer residual path around the propagation kernel, (ii) the linear projections inherited from attention, and (iii) the intermediate channel-extension projections that previously expanded channels before propagation. Cumulatively, these edits yield a $\sim\!5.5\times$ reduction in overhead latency (\cref{fig:overhead}).

\paragraph{Fused CUDA normalization.}
We further optimize the sigmoid-based row-stochastic normalization (\cref{eq:row_stochastic_norm}) by fusing its sequence of operations---sigmoid activation, local reduction, clamping, and division---into a single custom CUDA kernel. Executing all steps in one pass eliminates intermediate memory traffic and launch overhead, giving a $2.15\times$ speedup over PyTorch's baseline. Combined with the latent-space structural reduction ($C\!\to\!C_c$; e.g., $1152 \to 64$), the effective cost of normalization drops by $83.68\times$ at 1K resolution. Together, latent-space propagation, overhead removal, and fused normalization yield a $13.7\times$ speedup of the full GSPN layer at 1K; full comparisons appear in \cref{sec:experiment}.

\subsection{Training a New Architecture Fast: Cross-Operator Distillation}
\label{sec:distill}

\begin{figure}[t]
    \centering
    \includegraphics[width=\columnwidth]{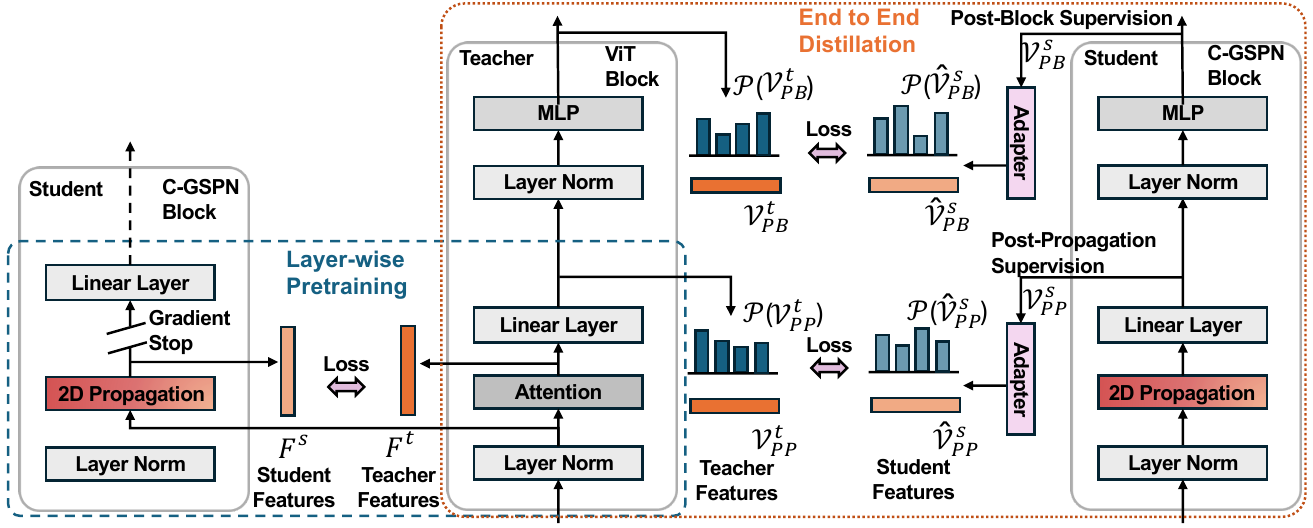}
    \caption{\textbf{Two-stage cross-operator distillation (Improvement~3).} \emph{Stage~1 (sublayer-wise)}: each C-GSPN propagation sublayer is aligned to the teacher's attention sublayer from the shared block input, giving a strong initialization. \emph{Stage~2 (end-to-end)}: the full student is distilled with two supervision taps per block---post-propagation (PP) and post-block (PB)---through lightweight feature adaptors that bridge the propagation/attention feature gap. This avoids training the new operator from scratch at foundation scale.}
    \label{fig:distill-loss}
\end{figure}

The efficient block of \cref{sec:efficiency} gives a fast \emph{forward} operator, but a foundation encoder still has to be \emph{trained}---and here the new architecture pays a hidden cost. Although GSPN~\citep{Wang2025GSPN} performs strongly on mid-scale tasks such as classification and generation, its use in foundation-scale vision models remains underexplored, and unlike a ViT student it \emph{does not inherit pretrained attention weights}: spatial propagation and attention are different operators, so the usual shortcut of warm-starting from a released checkpoint is generally unavailable. Training from scratch at CLIP/SigLIP scale (tens of billions of pairs) is very costly, so the question becomes \emph{how to train this new architecture efficiently}. We answer it by distilling from a pretrained quadratic-attention teacher into the GSPN-based student, aligning C-GSPN's block design with a SigLIP-2-style ViT~\citep{zhai2023sigmoid}. Cross-operator transfer is non-trivial: attention mixes tokens via explicit pairwise interactions, whereas GSPN attains global context through sequential local propagation that reduces the effective sequence length to $\sqrt{N}$. This mismatch makes direct attention-weight transfer inappropriate and induces a feature-distribution gap. We address it with a progressive, two-stage scheme (\cref{fig:distill-loss}) that makes foundation-scale training affordable.

\paragraph{Stage 1: Sublayer-wise pretraining.}
We first align each C-GSPN propagation sublayer with its corresponding teacher attention sublayer. For block $i$, both teacher and student take the output of the $(i{-}1)$-th teacher block as input:
\begin{equation}
\mathbf{h}^{t,(0)}=\mathbf{x}, \qquad \mathbf{h}^{t,(i)} = \mathrm{TeacherBlock}^{(i)}\big(\mathbf{h}^{t,(i-1)}\big).
\end{equation}
Given this shared input, we compute the sublayer features
\begin{equation}
F^{s,(i)} = f_{\text{C-GSPN-prop}}^{(i)}\big(\mathbf{h}^{t,(i-1)}\big), \qquad
F^{t,(i)} = f_{\text{Attention}}^{(i)}\big(\mathbf{h}^{t,(i-1)}\big),
\end{equation}
taken immediately after the student's propagation sublayer and the teacher's attention sublayer, and minimize a simple feature-alignment loss $\mathcal{L}^{(i)}_{\text{prop}} = \|F^{s,(i)} - F^{t,(i)}\|_2^2$. The teacher is frozen and gradients flow only through the student sublayer; each block is trained independently, without backpropagation across blocks, so every C-GSPN sublayer directly learns its paired attention sublayer's representation. This parallel scheme stabilizes training and provides a strong initialization for end-to-end distillation.

\paragraph{Stage 2: End-to-end distillation with dual taps.}
We then optimize end-to-end with \emph{two supervision taps per block}. We call the feature after the propagation/attention sublayer \emph{post-propagation} (PP) and the feature after the entire block (sublayer + MLP + norms) \emph{post-block} (PB). The rationale is to decompose cross-operator transfer: PB supervision preserves the teacher's block transformation, where the MLP is largely isomorphic across student and teacher, while PP supervision directly pressures the GSPN sublayer to reproduce the teacher's attention-style mixing rather than letting the MLP absorb the mismatch. Let $V^{s/t}_{\text{PP}}$ and $V^{s/t}_{\text{PB}}$ denote student/teacher features at the two taps and $P(\cdot)$ the token-wise softmax. We combine MSE feature alignment with KL distribution matching:
\begin{align}
\mathcal{L}_{\text{PP}}&=\mathrm{MSE}\!\big(\hat V^{s}_{\text{PP}},V^{t}_{\text{PP}}\big)+\lambda_{1}\,\mathrm{KL}\!\big(P(\hat V^{s}_{\text{PP}})\,\|\,P(V^{t}_{\text{PP}})\big), \\
\mathcal{L}_{\text{PB}}&=\mathrm{MSE}\!\big(\hat V^{s}_{\text{PB}},V^{t}_{\text{PB}}\big)+\lambda_{2}\,\mathrm{KL}\!\big(P(\hat V^{s}_{\text{PB}})\,\|\,P(V^{t}_{\text{PB}})\big),
\label{eq:distill_losses}
\end{align}
with total objective $\mathcal{L}_{\text{total}}=\alpha\,\mathcal{L}_{\text{PP}}+\beta\,\mathcal{L}_{\text{PB}}$.

\paragraph{Feature adaptors.}
Even with dual supervision, directly matching C-GSPN and ViT features is hard because the operators compute representations differently---propagation aggregates context sequentially while attention mixes all tokens at once---so raw-space comparison destabilizes training. We therefore insert lightweight \emph{feature adaptors} (2-layer MLPs) before each tap that map the student features into an aligned space ($V^{s}_{\text{PP}}\!\to\!\hat V^{s}_{\text{PP}}$, $V^{s}_{\text{PB}}\!\to\!\hat V^{s}_{\text{PB}}$ in \cref{eq:distill_losses}). By turning direct feature matching into learnable feature alignment, the adaptors stabilize optimization where the operator gap is largest (PP) and improve downstream accuracy (\cref{sec:ablation_study}). Similar tap-supervision and staged cross-architecture principles have been validated in recent distillation work~\citep{touvron2021training,yang2022vitkd,bick2024transformers}.

\paragraph{A small attention budget helps.}
Finally, inspired by hybrid Mamba--Transformer designs such as MaTVLM~\citep{li2025matvlm}, we find that retaining a modest fraction of attention layers gives a better accuracy--latency trade-off than either pure attention or pure C-GSPN. We adopt a hybrid that replaces every ninth GSPN block with a standard attention block (a $3/27$ ratio), injecting sparse pairwise mixing at regular depths while keeping the network subquadratic overall; we validate this in \cref{sec:ablation_study}.

\subsection{High-Resolution Encoder Distillation}
\label{sec:highres-transfer}

High-resolution downstream tasks are usually handled by tiling~\citep{liu2025nvila,liu2024llavanext} because attention scales quadratically with resolution, but tiling adds engineering complexity, boundary artifacts, and loss of global context. C-GSPN instead maintains low latency at 1K--2K and supports single-pass inference without tiling (\cref{sec:performance}); crucially, since it uses no positional embeddings, moving to higher resolutions requires no architectural change---only adapted training.

We study how to transfer low-resolution checkpoints to higher resolutions under limited compute, and find two challenges. First, naively transferring from a base resolution $r_0$ (e.g., $378$) to a target $r_K$ (e.g., $756$) performs poorly; a resolution curriculum~\citep{Qwen-VL,chen2023pali,chen2023pali-x,li2024monkey} that gradually increases resolution ($378\!\to\!518\!\to\!756$) substantially improves results ($80.4\%$ vs.\ $70.2\%$ under equal sample budgets). Second, contrastive supervision alone suffices for classification but fails to capture the fine-grained spatial detail needed for dense tasks. We therefore combine curriculum learning with \emph{upsampling self-distillation}: at each step $k>0$ the checkpoint from $r_{k-1}$ is frozen as a teacher, its post-propagation and post-block features are bilinearly upsampled to $r_k$, and they supervise the student at $r_k$ with the dual objectives of \cref{sec:distill}:
\begin{equation}
\tilde V_{m}^{t,(k)} = \mathrm{Up}\!\big(V_{m}^{t,(k-1)}\big), \quad
\tilde V_{b}^{t,(k)} = \mathrm{Up}\!\big(V_{b}^{t,(k-1)}\big), \quad
\mathcal{L}^{(k)}_{\mathrm{hr}} = \alpha\,\mathcal{L}^{(k)}_{\mathrm{module}} + \beta\,\mathcal{L}^{(k)}_{\mathrm{block}},
\end{equation}
where $\mathrm{Up}(\cdot)$ is bilinear upsampling from $r_{k-1}$ to $r_k$ and the per-tap losses follow the same MSE+KL form. Despite using approximate (upsampled) supervision, this staged self-distillation substantially improves dense-task performance while preserving C-GSPN's global context modeling (\cref{sec:experiment}), and requires only a small fraction of the data needed to train at high resolution from scratch.

\section{Experiments}
\label{sec:experiment}

Our evaluation mirrors the paper's two levels of efficiency and the end-to-end payoff. \textbf{Level 1:} we validate the fast GSPN kernel as a fast, accuracy-preserving primitive---a detailed profiling of the CUDA redesign (\cref{subsec:profiling}) followed by ImageNet classification and high-resolution text-to-image synthesis (\cref{sec:gspn2_vision})---confirming the operator is strong on its own. \textbf{Level 2:} we show the efficient block converts this kernel speed into block- and model-level speed (\cref{sec:performance}), and that cross-operator distillation yields a foundation-scale encoder with competitive zero-shot, segmentation, and detection quality at CLIP scale plus cheap high-resolution transfer (\cref{performance_over_vision_tasks}); ablations isolate the contribution of the block design and the training recipe (\cref{sec:ablation_study}). Unless noted, all latency is measured on A100 GPUs.

\subsection{Fast GSPN Kernel Efficiency}
\label{subsec:profiling}

We profile the fast GSPN kernel across input configurations, analyzing memory throughput, cache behavior, and SM utilization, and isolate the contribution of each optimization. Throughout this subsection, the baseline is the original GSPN reference kernel.

\paragraph{Step-by-step CUDA optimization.}
For a typical configuration ($1024\times1024$ image, batch size $16$, $8$ channels), \cref{fig:journey} quantifies each optimization term. The GSPN baseline takes $71.4$ ms due to launch overhead and poor memory access. A single fused kernel (\cref{subsec:single_kernel}) removes the micro-launches for a $1.2\times$ gain ($57.4$ ms); \textbf{coalesced memory access} (\cref{subsec:gspn2_cuda_opts}) maximizes bandwidth for a $23.9\times$ jump ($2.4$ ms); a \textbf{shared-memory cache} for hidden states adds $1.1\times$ ($2.2$ ms); \textbf{2D thread blocks} add $1.1\times$ ($2.1$ ms); and \textbf{compressive channels} (\cref{sec:gspnv2_theory}) add a final $1.1\times$ ($1.9$ ms). The fully optimized fast GSPN kernel achieves a $40.0\times$ cumulative speedup ($1.8$ ms). The relative impact of each term is workload-dependent; \cref{app:effablat_large_batch} analyzes an alternative large-batch configuration where coalesced access remains dominant but shared-memory caching and 2D blocks become configuration-sensitive.

\paragraph{Memory throughput.}
\cref{tab:kernel_throughput} reports Nsight Compute measurements: the fast GSPN kernel reaches near-theoretical global-memory throughput ($\sim$$93\%$ on A100), stable across batch sizes and resolutions, whereas the original GSPN kernel achieves only $3$--$8\%$ of peak and degrades as inputs grow.

\begin{table}[h]
\centering
\caption{\textbf{Global memory throughput under typical input configurations on A100.} Across input sizes, batch sizes, and channel counts representative of common deployments, the fast GSPN kernel consistently approaches peak bandwidth while the original GSPN kernel remains well below it.}
\label{tab:kernel_throughput}
\begin{tabular}{c|c|c|c|c}
\toprule
\rowcolor{ngreen!40}
\textbf{Input Size} & \textbf{Batch} & \textbf{Channels} & \textbf{GSPN (orig.)} & \textbf{fast GSPN} \\
\midrule
32$\times$32 & 32 & 196 & 114 GB/s (6.0\%) & 1832 GB/s (91.8\%) \\
64$\times$64 & 1 & 768 & 86 GB/s (4.5\%) & 1847 GB/s (92.3\%) \\
64$\times$64 & 1 & 1152 & 35 GB/s (2.1\%) & 1837 GB/s (92.0\%) \\
64$\times$64 & 1 & 32 & 125 GB/s (6.3\%) & 1830 GB/s (91.5\%) \\
128$\times$128 & 1 & 32 & 98 GB/s (4.9\%) & 1865 GB/s (93.3\%) \\
256$\times$256 & 1 & 64 & 76 GB/s (3.8\%) & 1842 GB/s (92.1\%) \\
256$\times$256 & 8 & 64 & 94 GB/s (4.7\%) & 1858 GB/s (92.9\%) \\
512$\times$512 & 1 & 128 & 64 GB/s (3.2\%) & 1840 GB/s (92.0\%) \\
\bottomrule
\end{tabular}
\end{table}

\begin{figure}[t]
    \centering
    \includegraphics[width=\textwidth]{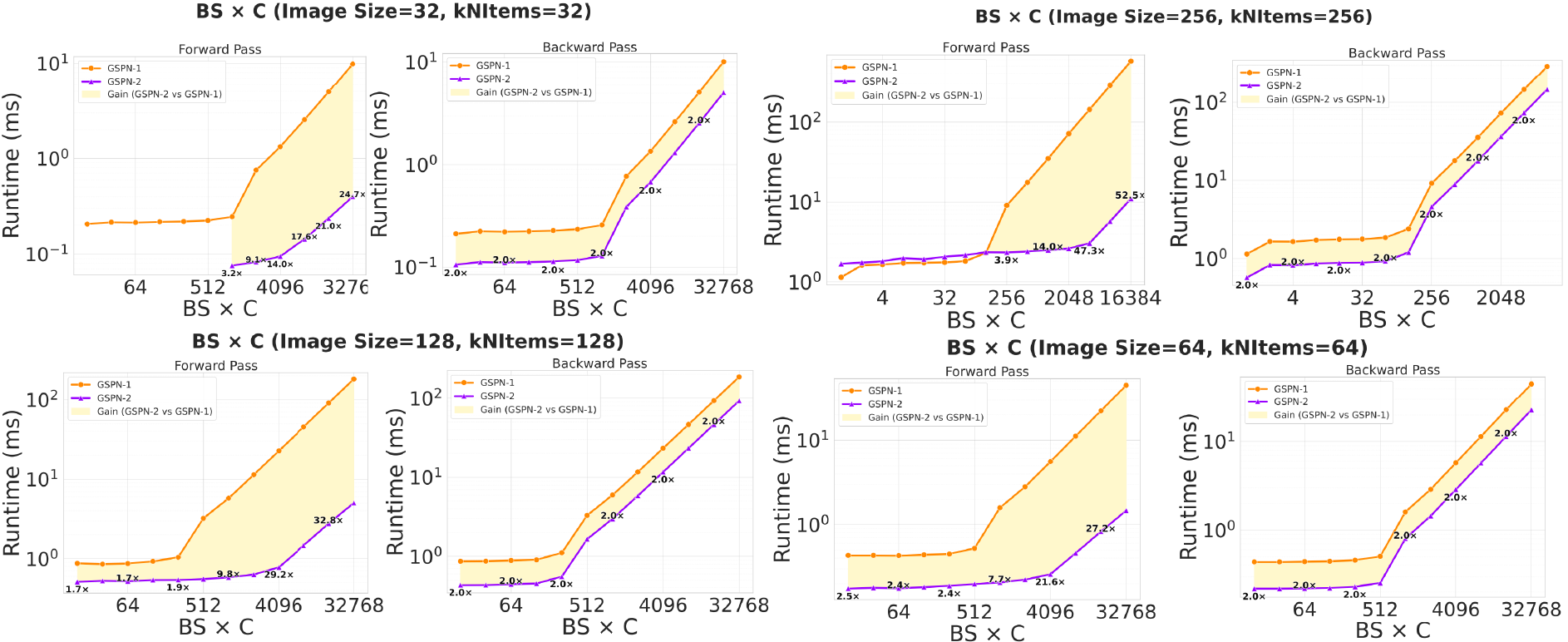}
    \caption{\textbf{Runtime comparison of the original GSPN kernel and the fast GSPN kernel.} Forward and backward execution times (ms) across channel counts and configurations. The fast GSPN kernel greatly improves both passes across cases.}
    \label{fig:runtime}
    \vspace{-0.5em}
\end{figure}

\paragraph{Scaling with input size, batch, and channels.}
\looseness=-1
\cref{fig:runtime} (upper row) shows the fast GSPN kernel consistently outperforms the original GSPN kernel across resolutions at fixed batch/channels, with up to $36.8\times$ (forward) and $25.3\times$ (backward) speedups at $1024\times1024$; the gap widens with resolution. The lower row shows the regime critical for video generation and foundation vision towers: even as batch sizes reach $256$ or channels reach $1024$, fast GSPN sustains $2$--$4\times$ speedups, e.g.\ a $27.4\times$ forward and $48.6\times$ backward speedup at $256$ channels, with the channel-sharing scheme (\cref{sec:gspnv2_theory}) adding up to $1.5\times$.

\paragraph{L1 cache and SM utilization.}
Profiling reveals that the L1 cache is highly effective for the fast GSPN kernel's structured, coalesced access: with explicit shared-memory caching, L1 hit rates drop to near $0\%$ (accesses served from shared memory) yet latency is comparable to relying on L1 (which shows $\sim$$35\%$ hit rates), so we keep shared memory for portability and determinism. SM occupancy approaches $100\%$ for large batch/channel workloads but can fall to $20$--$30\%$ for small inputs, where each independent chunk maps to a single block---pointing to further decomposition opportunities for low-workload regimes.

\subsection{Fast GSPN (Operator) on Vision Tasks}
\label{sec:gspn2_vision}

\begin{table*}[t]
\renewcommand{\arraystretch}{1.5}
    \caption{
    \textbf{ImageNet-1K performance at resolution $224^2$.} Colors denote backbone type: \textcolor{Yellow}{\textbf{yellow}} for CNNs, \textcolor{RPurple}{\textbf{orange}} for Transformers, and \textcolor{Green}{\textbf{green}} for raster-scan (1D linear propagation) methods; the line-scan models---prior GSPN and our fast GSPN---are highlighted in blue.
    }
    \label{tab:cls-imagenet}
    \centering
    \hspace*{-0.01\textwidth}%
    \begin{minipage}{.32\textwidth}
    \tiny
    \centering
    \setlength{\tabcolsep}{.1pt}
    \begin{tabular}{l | c | c | c c }
    \toprule
    \multirow{3}{*}{\makecell[c]{Model}}    & \multirow{3}{*}{\makecell[c]{Backbone}}     & \multirow{3}{*}{\makecell[c]{Param \\ (M)}}   & \multicolumn{2}{c}{IN-1K} \\
    \cline{4-5}
    ~ & ~ & ~ &  \multirow{2}{*}{\makecell[c]{MAC \\ (G)}} & \multirow{2}{*}{\makecell[c]{Acc \\ (\%)}} \\
    ~ & ~ & ~ & ~ & ~ \\
    \midrule
    \rowcolor{Yellow}
    ConvNeXT-T \cite{liu2022convnet} &  CN   & 29 & 4.5 & 82.1  \\
    \rowcolor{Yellow}
    MambaOut-Tiny \cite{yu2024mambaout} & CN & 27 & 4.5 & 82.7 \\
    \rowcolor{RPurple}
    DeiT-S \cite{touvron2021training} &  TF  & 22 & 4.6 & 79.8  \\
    \rowcolor{RPurple}
    T2T-ViT-14 \cite{yuan2021tokens} &  TF   & 22 & 4.8 & 81.5 \\
    \rowcolor{RPurple}
    Swin-T \cite{liu2021swin} &  TF   & 29 & 4.5 & 81.3 \\
    \rowcolor{RPurple}
    SwinV2-T \cite{liu2021swinv2} & TF & 28 & 4.4 & 81.8 \\
    \rowcolor{RPurple}
    CSWin-T \cite{dong2022cswin} &  TF  & 23 & 4.3 & 82.7  \\
    \rowcolor{RPurple}
    CoAtNet-0 \cite{dai2021coatnet} &   TF  & 25 & 4.2 & 81.6 \\
    \rowcolor{Green}
    Vim-S \cite{zhu2024ViM} & RS & 26 & 5.1 & 80.5 \\
    \rowcolor{Green}
    VMamba-T \cite{liu2024vmamba} &  RS & 22 & 5.6 & 82.2 \\
    \rowcolor{Green}
    Mamba-2D-S \cite{li2024mamba} &  RS & 24 & -- & 81.7 \\
    \rowcolor{Green}
    LocalVMamba-T \cite{huang2024localmamba} & RS & 26 & 5.7 & 82.7 \\
    \rowcolor{Green}
    VRWKV-S \cite{duan2024visionrwkv} & RS & 24 & 4.6 & 80.1 \\
    \rowcolor{Green}
    ViL-S \cite{alkin2024visionlstm} & RS & 23 & 5.1 & 81.5 \\
    \rowcolor{Green}
    MambaVision-T ~\citetalias{hatamizadeh2024mambavision} & RS & 32 & 4.4 & 82.3 \\
    \midrule
    \rowcolor{Blue}
    GSPN-T & Line & 30 & 5.3 & \textbf{83.0} \\
    \rowcolor{Blue}
    \textbf{Fast GSPN-T (Ours)} & Line & 24 & 4.2 & \textbf{83.0} \\
    \bottomrule
    \end{tabular}
    \end{minipage}
\renewcommand{\arraystretch}{1.3}
    \begin{minipage}{.32\textwidth}
    \tiny
    \centering
    \setlength{\tabcolsep}{.1pt}
    \begin{tabular}{l | c | c | c c }
    \toprule
    \multirow{3}{*}{\makecell[c]{Model}}    & \multirow{3}{*}{\makecell[c]{Backbone}}     & \multirow{3}{*}{\makecell[c]{Param \\ (M)}}   & \multicolumn{2}{c}{IN-1K} \\
    \cline{4-5}
    ~ & ~ & ~ &  \multirow{2}{*}{\makecell[c]{MAC \\ (G)}} & \multirow{2}{*}{\makecell[c]{Acc \\ (\%)}} \\
    ~ & ~ & ~ & ~ & ~ \\
    \midrule
    \rowcolor{Yellow}
    ConvNeXT-S \cite{liu2022convnet} & CN & 50 & 8.7 & 83.1 \\
    \rowcolor{Yellow}
    CNFormer-S36 \cite{yu2024metaformer} & CN & 40 & 7.6 & 84.1 \\
    \rowcolor{Yellow}
    MogaNet-B \cite{li2024MogaNet} & CN & 44 & 9.9 & 84.3 \\
    \rowcolor{Yellow}
    InternImage-S \cite{wang2023internimage} & CN & 50 & 8.0 & 84.2 \\
    \rowcolor{Yellow}
    MambaOut-Small \cite{yu2024mambaout} & CN & 48 & 9.0 & 84.1 \\
    \rowcolor{RPurple}
    T2T-ViT-19 \cite{yuan2021tokens} & TF & 39 & 8.5 & 81.9 \\
    \rowcolor{RPurple}
    Focal-Small \cite{yang2022focal} & TF & 51 & 9.1 & 83.5 \\
    \rowcolor{RPurple}
    BiFormer-B \cite{zhu2023biformer} & TF & 57 & 9.8 & 84.3 \\
    \rowcolor{RPurple}
    NextViT-B \cite{li2022next} & TF & 45 & 8.3 & 83.2 \\
    \rowcolor{RPurple}
    Twins-B \cite{chu2021twins} & TF & 56 & 8.3 & 83.1 \\
    \rowcolor{RPurple}
    MaxViT-Small \cite{tu2022maxvit} & TF & 69 & 11.7 & \textbf{84.4} \\
    \rowcolor{RPurple}
    Swin-S \cite{liu2021swin} & TF & 50 & 8.7 & 83.0 \\
    \rowcolor{RPurple}
    SwinV2-S \cite{liu2021swinv2} & TF &  50 & 8.5 & 83.8 \\
    \rowcolor{RPurple}
    CoAtNet-1 \cite{dai2021coatnet} &  TF & 42 & 8.4 & 83.3 \\
    \rowcolor{RPurple}
    UniFormer-B \cite{li2022uniformer} &  TF & 50 & 8.3 & 83.9 \\
    \rowcolor{Green}
    VMamba-S \cite{liu2024vmamba} &  RS & 44 & 11.2 & 83.5 \\
    \rowcolor{Green}
    LocalVMamba-S \cite{huang2024localmamba} &  RS & 50 & 11.4 & 83.7 \\
    \rowcolor{Green}
    MambaVision-S ~\citetalias{hatamizadeh2024mambavision} & RS & 50 & 7.5 & 83.3 \\
    \midrule
    \rowcolor{Blue}
    GSPN-S & Line & 50 & 9.0 & 83.8 \\
    \rowcolor{Blue}
    \textbf{Fast GSPN-S (Ours)} & Line & 50 & 9.2 & \textbf{84.4} \\
    \bottomrule
    \end{tabular}
    \end{minipage}
\renewcommand{\arraystretch}{1.35}
    \begin{minipage}{.32\textwidth}
    \tiny
    \centering
    \setlength{\tabcolsep}{.1pt}
    \begin{tabular}{l | c | c | c c }
    \toprule
    \multirow{3}{*}{\makecell[c]{Model}}    & \multirow{3}{*}{\makecell[c]{Backbone}}     & \multirow{3}{*}{\makecell[c]{Param \\ (M)}}   & \multicolumn{2}{c}{IN-1K} \\
    \cline{4-5}
    ~ & ~ & ~ &  \multirow{2}{*}{\makecell[c]{MAC \\ (G)}} & \multirow{2}{*}{\makecell[c]{Acc \\ (\%)}} \\
    ~ & ~ & ~ & ~ & ~ \\
    \midrule
    \rowcolor{Yellow}
    ConvNeXT-B \cite{liu2022convnet} & CN & 89 & 15.4 & 83.8 \\
    \rowcolor{Yellow}
    CNFormer-M36 \cite{yu2024metaformer} & CN & 57 & 12.8 & 84.5 \\
    \rowcolor{Yellow}
    MambaOut-Base \cite{yu2024mambaout} & CN & 85 & 15.8 & 84.2 \\
    \rowcolor{Yellow}
    SLaK-B \cite{liu2023more} & CN & 95 & 17.1 & 84.0 \\
    \rowcolor{RPurple}
    DeiT-B \cite{touvron2021training} & TF & 86 & 17.5 & 81.8 \\
    \rowcolor{RPurple}
    T2T-ViT-24 \cite{yuan2021tokens} & TF & 64 & 13.8 & 82.3 \\
    \rowcolor{RPurple}
    Swin-B \cite{liu2021swin} & TF & 88 & 15.4 & 83.5 \\
    \rowcolor{RPurple}
    SwinV2-B \cite{liu2021swinv2} & TF & 88 & 15.1 & \textbf{84.6} \\
    \rowcolor{RPurple}
    CSwin-B \cite{dong2022cswin} & TF & 78 & 15.0 & 84.2 \\
    \rowcolor{RPurple}
    MViTv2-B \cite{li2021improved} & TF & 52 & 10.2 & 84.4 \\
    \rowcolor{RPurple}
    CoAtNet-2 \cite{dai2021coatnet} &  TF & 75 & 15.7 & 84.1 \\
    \rowcolor{Green}
    Vim-B \cite{zhu2024ViM} & RS & 98 & 17.5 & 81.9 \\
    \rowcolor{Green}
    VMamba-B \cite{liu2024vmamba} &  RS & 89 & 15.4 & 83.9 \\
    \rowcolor{Green}
    Mamba-2D-B \cite{li2024mamba} &  RS & 92 & -- & 83.0 \\
    \rowcolor{Green}
    VRWKV-B \cite{duan2024visionrwkv} & RS & 94 & 18.2 & 82.0 \\
    \rowcolor{Green}
    ViL-B \cite{alkin2024visionlstm} & RS & 89 & 18.6 & 82.4 \\
    \rowcolor{Green}
    MambaVision-B ~\citetalias{hatamizadeh2024mambavision} & RS & 98 & 15.0 & 84.2 \\
    \midrule
    \rowcolor{Blue}
    GSPN-B & Line & 89 & 15.9 & 84.3 \\
    \rowcolor{Blue}
    \textbf{Fast GSPN-B (Ours)} & Line & 89 & 14.2 & \textbf{84.9} \\
    \bottomrule
    \end{tabular}
    \end{minipage}
\end{table*}

\paragraph{Image classification.}
\cref{tab:cls-imagenet} compares ImageNet-1K accuracy across ConvNet, Transformer, and sequential (raster-scan) backbones. Here we evaluate fast GSPN as a classification \emph{operator}/backbone (the fast GSPN kernel with its compact channel propagation), denoting its variants fast GSPN-T/S/B; the prior GSPN backbones (GSPN-T/S/B) are the baseline. For fast GSPN we share $w_i$ across channels in all modules, set the compressive proxy dimension $C_{\text{proxy}}{=}2$ (reallocating saved parameters to depth/width), add a Local Perception Unit at the start of each block and FFN, and apply MESA to curb overfitting. Fast GSPN-T reaches $83.0\%$ with fewer parameters (24M vs.\ 30M) and lower cost (4.2G vs.\ 5.3G MACs) than GSPN-T, surpassing Vim-S ($80.5\%$), VMamba-T ($82.2\%$), and LocalVMamba-T ($82.7\%$). Fast GSPN-S reaches $84.4\%$ ($+0.6\%$ over GSPN-S at equal parameters), ahead of MambaOut-Small ($84.1\%$) and UniFormer-B ($83.9\%$); fast GSPN-B reaches $84.9\%$ ($+0.6\%$ over GSPN-B) while reducing MACs (14.2G vs.\ 15.9G).

\begin{figure}[t]
    \centering
    \includegraphics[width=\textwidth]{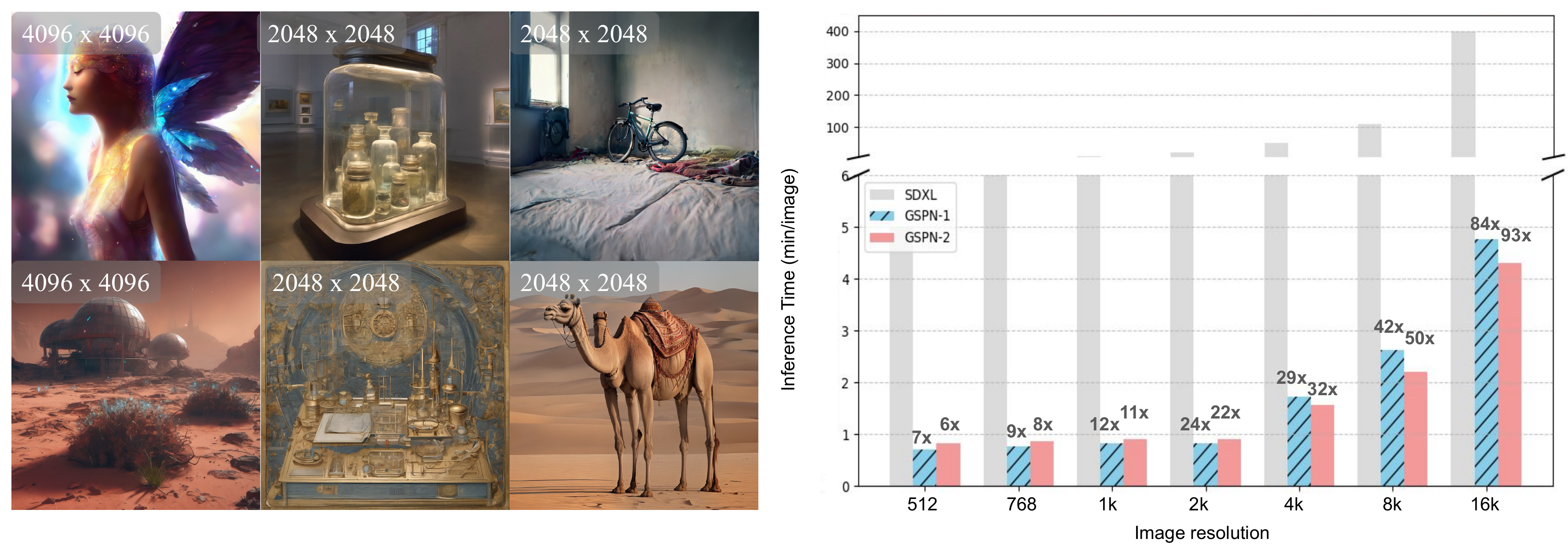}
    \caption{\textbf{Qualitative text-to-image results from our fast GSPN SDXL model.} We enable generation up to 16K resolution on a single A100 while reducing inference time by up to $93\times$.}
    \label{fig:vis1}
\end{figure}

\paragraph{Text-to-image generation.}
Integrated into Stable Diffusion XL with proxy compression to $1/8$ of channels ($C_{\text{proxy}}{=}C/8$), fast GSPN accelerates high-resolution synthesis without degrading quality (\cref{fig:vis1}). It achieves a $32\times$ speedup over the SDXL baseline at 4K, and at 16K reduces inference time by $93\times$ relative to the original GSPN's $84\times$ improvement, enabling 16K generation on a single A100.

\subsection{C-GSPN System Efficiency}
\label{sec:performance}

\begin{figure}[t]
    \centering
    \includegraphics[width=0.92\columnwidth]{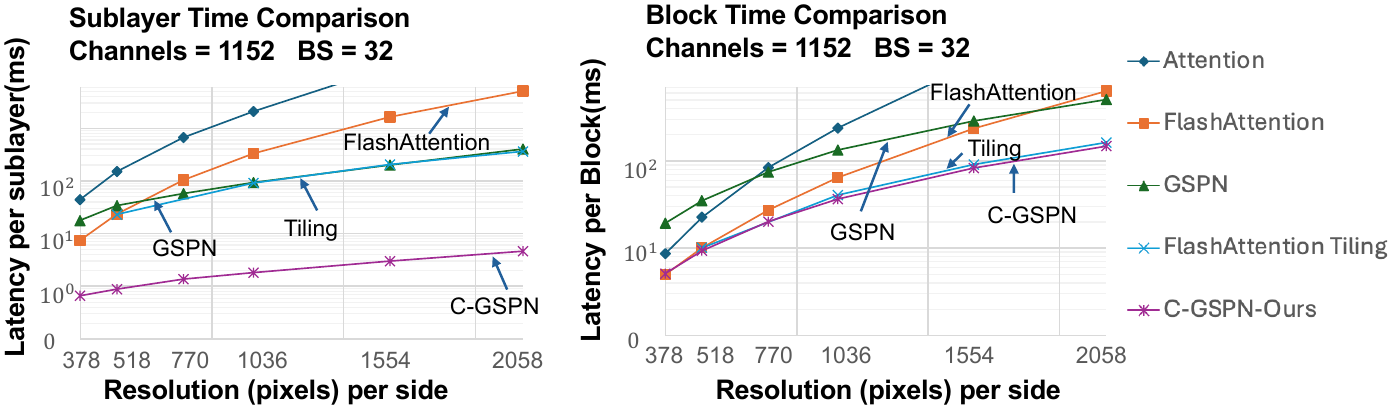}
    \caption{\textbf{Latency vs.\ resolution.} Sublayer (left) and full-block (right) latency for attention, FlashAttention, the original GSPN, and C-GSPN. Attention-based cores scale quadratically and quickly become memory- or latency-bound, while C-GSPN's latent-space propagation stays low and resolution-stable, translating its kernel-level gains (Improvement~1) into block-level gains (Improvement~2) at 1K--2K.}
    \label{fig:latency_profile}
\end{figure}

We benchmark C-GSPN against standard attention, FlashAttention, and the original GSPN at the propagation \emph{sublayer}, the full \emph{layer}, the complete transformer \emph{block} (core + MLP + norms + residuals), and end-to-end throughput, at batch size $32$ and $1152$ channels (\cref{tab:latency_breakdown}; more resolutions in \cref{app:more_latency}). \cref{fig:latency_profile} summarizes how these costs scale with resolution.

\begin{table*}[t]
    \centering
    \resizebox{\textwidth}{!}{%
    \begin{tabular}{l l l l l l l l l l l l l l l l l}
        \toprule
        & \multicolumn{4}{c}{\textbf{Sublayer latency (ms)}} & \multicolumn{4}{c}{\textbf{Layer latency (ms)}} & \multicolumn{4}{c}{\textbf{Block latency (ms)}} & \multicolumn{4}{c}{\textbf{Throughput (img/s)}} \\
        \cmidrule(lr){2-5} \cmidrule(lr){6-9} \cmidrule(lr){10-13} \cmidrule(lr){14-17}
        & \multicolumn{2}{c}{\textbf{1036}} & \multicolumn{2}{c}{\textbf{2058}} & \multicolumn{2}{c}{\textbf{1036}} & \multicolumn{2}{c}{\textbf{2058}} & \multicolumn{2}{c}{\textbf{1036}} & \multicolumn{2}{c}{\textbf{2058}} & \multicolumn{2}{c}{\textbf{1036}} & \multicolumn{2}{c}{\textbf{2058}} \\
        \cmidrule(lr){2-3} \cmidrule(lr){4-5} \cmidrule(lr){6-7} \cmidrule(lr){8-9} \cmidrule(lr){10-11} \cmidrule(lr){12-13} \cmidrule(lr){14-15} \cmidrule(lr){16-17}
        & \textbf{ms} & \textbf{$\times$} & \textbf{ms} & \textbf{$\times$} & \textbf{ms} & \textbf{$\times$} & \textbf{ms} & \textbf{$\times$} & \textbf{ms} & \textbf{$\times$} & \textbf{ms} & \textbf{$\times$} & \textbf{img/s} & \textbf{$\times$} & \textbf{img/s} & \textbf{$\times$} \\
        \midrule
        Attention      & 205.90 & \cellcolor[HTML]{F1FAF1}1143.89$\times$ & OOM & \cellcolor[HTML]{F1FAF1}OOM & 218.22 & \cellcolor[HTML]{F1FAF1}13.11$\times$ & OOM & \cellcolor[HTML]{F1FAF1}OOM & 238.18 & \cellcolor[HTML]{F1FAF1}6.51$\times$ & OOM & \cellcolor[HTML]{F1FAF1}OOM & 4.83 & \cellcolor[HTML]{F1FAF1}5.85$\times$ & OOM & \cellcolor[HTML]{F1FAF1}OOM \\
        FlashAttention & 32.81 & \cellcolor[HTML]{F1FAF1}182.28$\times$ & 504.73 & \cellcolor[HTML]{F1FAF1}1097.24$\times$ & 44.31 & \cellcolor[HTML]{F1FAF1}2.66$\times$ & 550.87 & \cellcolor[HTML]{F1FAF1}8.25$\times$ & 68.93 & \cellcolor[HTML]{F1FAF1}1.88$\times$ & 631.63 & \cellcolor[HTML]{F1FAF1}4.28$\times$ & 16.89 & \cellcolor[HTML]{F1FAF1}1.67$\times$ & 1.81 & \cellcolor[HTML]{F1FAF1}3.82$\times$ \\
        GSPN           & 9.95 & \cellcolor[HTML]{F1FAF1}55.28$\times$ & OOM & \cellcolor[HTML]{F1FAF1}OOM & 113.16 & \cellcolor[HTML]{F1FAF1}6.80$\times$ & OOM & \cellcolor[HTML]{F1FAF1}OOM & 133.12 & \cellcolor[HTML]{F1FAF1}3.64$\times$ & OOM & \cellcolor[HTML]{F1FAF1}OOM & 6.53 & \cellcolor[HTML]{F1FAF1}4.33$\times$ & OOM & \cellcolor[HTML]{F1FAF1}OOM \\
        \rowcolor[HTML]{D0E8D0}\textbf{C-GSPN (ours)}  & \textbf{0.18} & \textbf{1$\times$} & \textbf{0.46} & \textbf{1$\times$} & \textbf{16.64} & \textbf{1$\times$} & \textbf{66.77} & \textbf{1$\times$} & \textbf{36.60} & \textbf{1$\times$} & \textbf{147.52} & \textbf{1$\times$} & \textbf{28.28} & \textbf{1$\times$} & \textbf{6.91} & \textbf{1$\times$} \\
        \bottomrule
    \end{tabular}%
    }
    \caption{\textbf{Latency breakdown and throughput across resolutions} (batch size $32$). Sublayer: latent-space 2D propagation vs.\ scaled dot-product attention. Layer: a C-GSPN layer vs.\ a multi-head attention layer. Block: end-to-end transformer block. Right: block throughput (img/s). {\setlength{\fboxsep}{1pt}\colorbox[HTML]{F1FAF1}{\strut Shaded columns}} report how much faster C-GSPN is than each baseline. The propagation core is up to $1097\times$ faster than dot-product attention at the sublayer and delivers a $4.28\times$ end-to-end block speedup; throughput improves $1.67\times$ and $3.82\times$ over FlashAttention at $1036$ and $2058$.}
    \label{tab:latency_breakdown}
\end{table*}

\paragraph{Sublayer and block results.}
The original GSPN remains $55.3\times$--$86.9\times$ slower than C-GSPN at 1K--2K. FlashAttention needs over $500$ ms per layer at 2K while C-GSPN's sublayer stays at $0.46$ ms---a $\sim$$1000\times$ gap. At the full block (including MLP, norms, residuals), FlashAttention exceeds $600$ ms at 2K whereas C-GSPN completes in under $150$ ms, a $4\times$ end-to-end speedup, with single-pass inference and no tiling.

\subsection{C-GSPN at Foundation Scale}
\label{performance_over_vision_tasks}

We distill C-GSPN from a strong attention teacher (OpenCLIP ViT-SO/14 at $378$) using only $600$M image--text pairs---far less than training attention models from scratch (e.g., SigLIP-v2's $40$B). Under identical data/compute we compare three students: an \emph{isomorphic} ViT$\to$ViT (identical architecture), an \emph{original} GSPN student, and our C-GSPN. We evaluate zero-shot classification (ImageNet Top-1/Top-5)~\citep{imagenet_lsvrc}, segmentation (ADE20K-F, ADE20K, PASCAL)~\citep{zhou2019semantic,everingham2010pascal}, and detection (COCO)~\citep{lin2014microsoft}.

\begin{table}[t]
    \centering
    \resizebox{0.7\columnwidth}{!}{%
    \begin{tabular}{l c  c c c | c}
        \toprule
        \textbf{Resolution} & \textbf{378} & \textbf{518} & \textbf{756} & \textbf{1036} & \textbf{Latency(1K)}\\
        \midrule
        ViT-Distill & 45.5 & -- & -- & 44.1 &  765.9(s)\\
        \midrule
        C-GSPN w/o KD & 46.0  &  45.1  & 44.5  & 43.5 &  \cellcolor[HTML]{F1FAF1} 319.4(s)\\
        C-GSPN w/ KD & 46.0  & 46.3  & 46.2  & 45.8 & \cellcolor[HTML]{F1FAF1} (2.40$\times$ Speed up) \\
        \bottomrule
    \end{tabular}}
    \caption{\textbf{High-resolution transfer under limited compute.} Segmentation accuracy (ADE20K) across increasing input resolutions; KD denotes our upsampling self-distillation. We also report eight-GPU parallel training latency at $1036$ per $1000$ iterations (batch size $1$): C-GSPN yields a $2.40\times$ speedup.}
    \label{tab:highrestransfer}
\end{table}

\begin{table*}[t]
    \centering
    \resizebox{1.0\textwidth}{!}{%
    \begin{tabular}{lcccccccccc}
    \toprule
    \multirow{2}{*}{\textbf{Method}} & \multirow{2}{*}{\textbf{Params.}} & \multirow{2}{*}{\textbf{Res.}} & \multirow{2}{*}{\textbf{Patches}} & \multicolumn{2}{c}{\textbf{Classification}} & \multicolumn{3}{c}{\textbf{Segmentation}} & \textbf{Detection} & \multirow{2}{*}{\textbf{Avg.}} \\
    \cmidrule(lr){5-6} \cmidrule(lr){7-9}
    & & & & \textbf{Top-1} & \textbf{Top-5} & \textbf{ADE20K-F} & \textbf{ADE20K} & \textbf{Pascal} & \textbf{COCO} & \\
     \midrule
\rowcolor[HTML]{F1F5FB} \textcolor{gray}{OpenCLIP SO/14 (teacher)} & \textcolor{gray}{427M} & \textcolor{gray}{378} & \textcolor{gray}{729} & \textcolor{gray}{84.1} & \textcolor{gray}{97.4} & \textcolor{gray}{42.8} & \textcolor{gray}{45.8} & \textcolor{gray}{77.5} & \textcolor{gray}{47.7} & \textcolor{gray}{64.6} \\
    \midrule
    \rowcolor[HTML]{F1FAF1} ViT-Distill & 427M & 378& 729&\textbf{82.2} &\textbf{96.7} &43.2 & \underline{45.5}& \underline{77.2}& \textbf{45.8} & \textbf{63.5} \\
    \rowcolor[HTML]{F1FAF1} GSPN & 477M & 378& 729&80.5 &95.8 &\underline{44.3} &45.3&\underline{77.2}& 44.3& 62.7 \\
    \rowcolor[HTML]{D0E8D0} \textbf{C-GSPN (ours)} & 365M & 378& 729& \underline{81.3}& \underline{96.3}& \textbf{44.7}& \textbf{46.0}&\textbf{77.6} &\underline{45.0} & \underline{63.3} \\
    \bottomrule
    \end{tabular}%
    }
    \caption{\textbf{Comprehensive evaluation across vision tasks.} OpenCLIP SO/14 is the teacher for all distilled models. Avg.\ is a macro average: mean(mean(Top-1, Top-5), mean(ADE20K-F, ADE20K, Pascal), COCO). ADE20K-F uses feature tokens as in EfficientViT~\citep{cai2023efficientvit}; ADE20K uses feature and summary tokens as in TIPS~\citep{tips_paper}.}
    \label{tab:comprehensive_comparison}
\end{table*}

\paragraph{Foundation-scale quality.}
Despite $15\%$ fewer parameters, C-GSPN matches the isomorphic ViT$\to$ViT baseline ($63.3$ vs.\ $63.5$ macro average), outperforms the original GSPN ($62.7$), and even exceeds the teacher on segmentation ($+0.2\%$ ADE20K) (\cref{tab:comprehensive_comparison}). To our knowledge, this is among the first demonstrations that a subquadratic spatial operator can be scaled to CLIP-level pretraining while retaining competitive quality. Combined with \cref{sec:performance}, C-GSPN is a viable efficient alternative: $4.28\times$ block-level speedup at high resolution while preserving task quality and enabling tiling-free single-pass inference.

\paragraph{High-resolution transfer.}
Using a lightweight resolution curriculum of $3$M samples ($1$M per stage, $378\!\to\!518\!\to\!756\!\to\!1036$) instead of full-scale training, the staged upsampling self-distillation of \cref{sec:highres-transfer} improves ADE20K segmentation at $518$ by $+1.2$ points over contrastive-only training and, at $1036$, reaches a $2.40\times$ training speedup over ViT-Distill (\cref{tab:highrestransfer}).

\subsection{Ablation Studies}
\label{sec:ablation_study}

\begin{figure*}[ht!]
    \centering
    \includegraphics[width=0.98\textwidth]{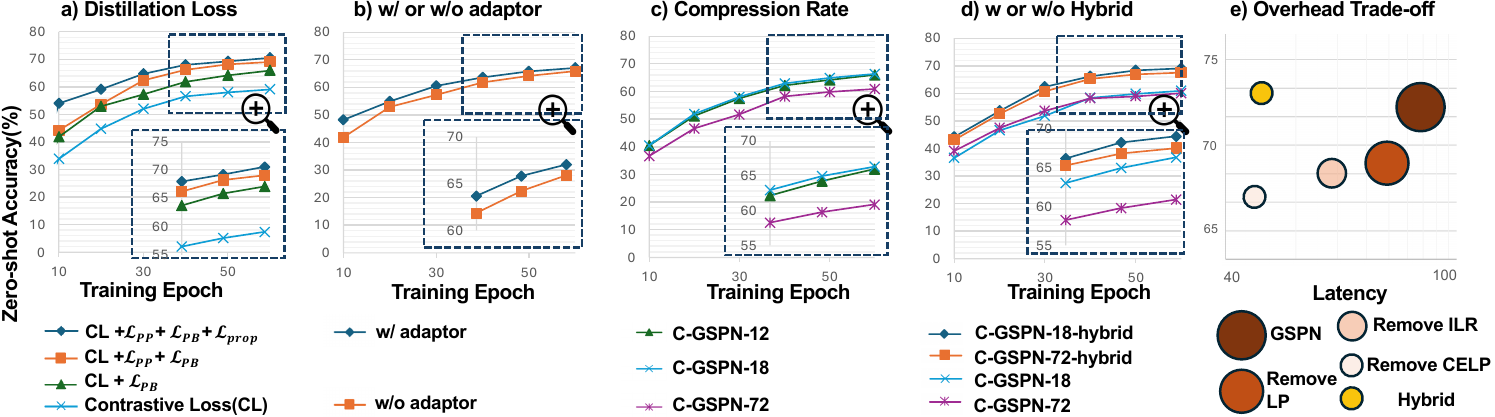}
    \caption{\textbf{Ablations on training strategy and module structure.} (a) Distillation: contrastive $\rightarrow$ +PB $\rightarrow$ +PB+PP (+adaptors); PP gives the largest gain, adaptors help, Stage-1 gives a strong start. (b) Adaptors consistently help. (c) Compression: among tested ratios, $18$ gives the best observed performance--accuracy balance. (d) Hybrid: adding $3/27$ attention layers improves accuracy while preserving speed. (e) Overhead trade-off: accuracy vs.\ latency at 1K; circle area encodes parameter count, and the $1/9$-attention hybrid lies on a favorable cost--quality frontier.}
    \label{fig:ablation}
\end{figure*}

We ablate the distillation stages and tap positions, feature adaptors, compression rate, and the attention budget (\cref{fig:ablation}).

\paragraph{(a) Distillation strategy.}
\emph{Stage-1} (sublayer-wise) aligns each propagation sublayer with its paired attention sublayer so the projections and MLP cannot absorb the operator gap, providing a strong initialization that persists through end-to-end training. In \emph{Stage-2}, contrastive loss alone underperforms under limited compute; prior feature-distillation methods supervise only post-block features, but adding direct supervision on the raw propagation output (PP) before the MLP gives the sublayer a dedicated signal, improving accuracy by $3.1\%$.

\paragraph{(b) Adaptors.}
Lightweight MLP adaptors before each tap bridge the feature spaces and yield consistent gains across epochs.

\paragraph{(c) Compression rate.}
Comparing ratios $12/18/72$, lower compression (more latent channels) improves capacity up to a point; reducing compression further yields little additional gain. Among these, ratio $18$ gives the best accuracy--efficiency balance.

\paragraph{(d) Attention budget.}
Replacing $3/27$ layers with attention consistently improves over pure C-GSPN: a small attention budget injects long-range pairwise mixing in a few layers while the rest retain efficient global propagation, avoiding quadratic cost throughout---a trend echoed in recent hybrid work~\citep{waleffe2024empirical,basant2025nvidia}.

\paragraph{(e) Overhead trade-off.}
Progressively removing the original GSPN's non-propagation components---additional linear projections (LP), the inner-layer residual (ILR), and channel-extension projections (CELP)---cuts overhead by $\sim$$5.5\times$ at 1K (\cref{fig:overhead}). Under an identical recipe, removing LP is nearly neutral, ILR incurs a small drop, and CELP causes the largest degradation. To recover capacity without losing the speedups, the $1/9$-attention hybrid restores most accuracy while staying $\approx\!3.9\times$ faster at 2K than pure-attention baselines and exceeding the GSPN variant on all tasks, making it our best cost--quality operating point.

\section{Limitations and Future Work}
\label{sec:limitations}

Our design has clear boundaries that suggest concrete next steps. At the \emph{kernel} level, the fast GSPN kernel's gains shrink when the product of batch size and channel count ($B\times C$) is small, since few resident blocks leave SMs underutilized (\cref{app:perf_batch_channel}); practical evaluation on long-context video remains underexplored, and the current operator lacks the CLS and register tokens that some ViT-based pipelines rely on, limiting drop-in use in models built around summary tokens. At the \emph{encoder} level, C-GSPN accelerates the propagation sublayer so effectively that the feed-forward MLP becomes the dominant cost: profiling at batch size $32$ shows the MLP accounts for $52\%$ of block latency at resolutions $\ge 512$. Our dense-prediction evaluations also focus on $378$--$1036$ resolutions; pushing further would better expose the scaling advantage of a subquadratic core. Future work will target MLP compression, kernel fusion, and low-rank feed-forward variants, integrate summary/register tokens into the propagation block, and extend C-GSPN to video and multimodal long-context settings, where tiling-free, linear-time spatial propagation should be most valuable.

\section{Conclusion}
\label{sec:conclusion}

We set out to turn 2D spatial propagation---a subquadratic operator that, unlike most efficient-attention variants, preserves the spatial structure of images---into a practical foundation-scale vision encoder. The central lesson is that this requires winning efficiency at \emph{two levels}, and that the first is necessary but not sufficient for the second. At the \textbf{system level}, the fast GSPN kernel reworks the line scan into a single warp-specialized CUDA kernel with compact, channel-shared propagation and careful shared-memory, coalescing, and stream concurrency, delivering up to $52\times$ speedups over the original GSPN kernel at near-peak bandwidth and a strong standalone operator. But kernel speed alone leaves block-level overhead and, crucially, the from-scratch training cost of a non-inheritable operator unaddressed. \textbf{C-GSPN} closes this gap at the \textbf{architecture and training levels}: a compressed latent-space block turns the fast scan into a fast \emph{block}, and a two-stage, dual-tap cross-operator distillation recipe makes foundation-scale training affordable without inheritable attention weights. The result is, to our knowledge, among the first subquadratic spatial operators scaled to CLIP-level pretraining---matching an isomorphic ViT with $15\%$ fewer parameters, improving dense prediction, and enabling tiling-free high-resolution inference with $2$--$4\times$ block-level speedups. Together, the two levels of efficiency help position 2D spatial propagation as a practical, hardware-efficient, and scalable basis for future vision foundation encoders.

\clearpage
\appendix
\renewcommand{\thefigure}{S\arabic{figure}}
\setcounter{figure}{0}
\renewcommand{\thetable}{S\arabic{table}}
\setcounter{table}{0}

\section{GPU Hardware and Kernel Execution for 2D Linear Propagation}
\label{app:gpu}

Modern GPUs such as NVIDIA's A100 expose parallelism through a hierarchical execution model of grids, thread blocks, and warps. A kernel is launched as a grid of thread blocks, each holding up to $1024$ threads organized into $32$-thread warps, the basic scheduling unit on the streaming multiprocessors (SMs; $108$ on the A100). Warps execute in single-instruction, multiple-thread (SIMT) fashion, maximizing throughput when occupancy---the fraction of active warps per SM---is high, balanced against register usage (up to $65{,}536$ per SM) and shared memory (up to $164$\,KB per SM).

In 2D line-scan propagation~\citep{Wang2025GSPN, liu2017learning}, an input tensor of shape $B\times C\times H\times W$ is processed by sequential row or column updates with parallel computation within each step. The CUDA implementation maps spatial positions to threads while $B$ and $C$ define independent slices for concurrent processing; a 1D block configuration might use a fixed \texttt{blockDim.x} (e.g., $512$) with grid size scaled by $B\times C\times H$. This faces scalability limits with large $B\times C$: each SM hosts at most $32$ resident blocks, so once $B\times C$ exceeds the device capacity, excess slices serialize and runtime spikes despite the theoretical parallelism. C-GSPN---through its fast GSPN kernel (\cref{sec:gspn2}) and compressed block (\cref{sec:cgspn})---is designed precisely to keep the active-block count within this regime.

\section{Comprehensive Fast GSPN Comparison on ImageNet-1K}
\label{app:gspn2_imagenet}

\begin{figure}[t]
    \centering
    \includegraphics[width=0.8\textwidth]{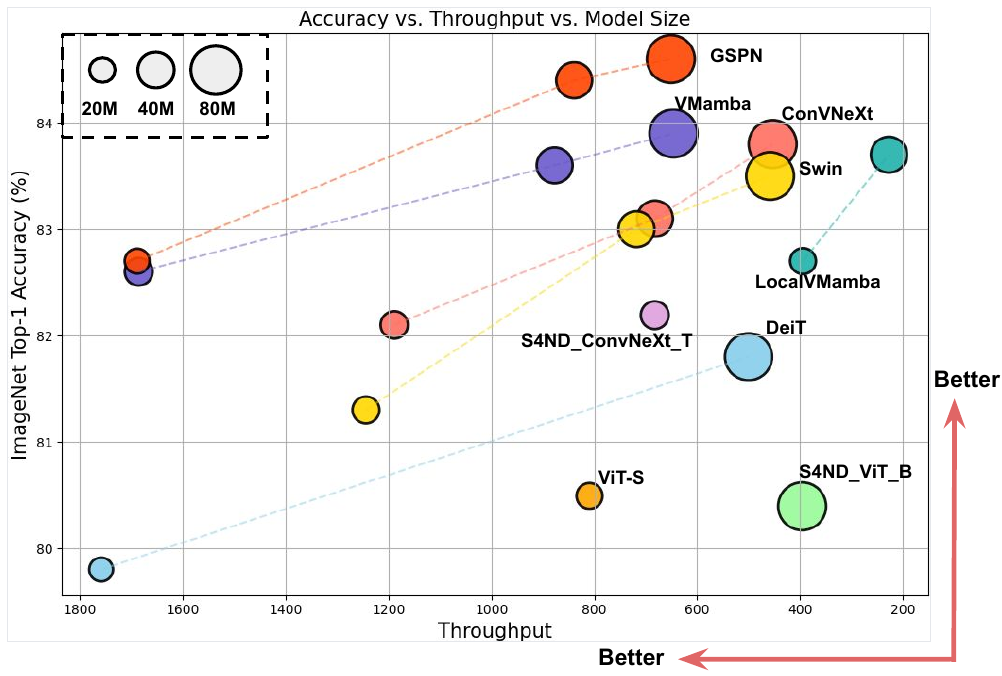}
    \caption{\textbf{Fast GSPN vs.\ state-of-the-art architectures on ImageNet-1K}, analyzing the trade-off between accuracy, model size, and throughput. Fast GSPN is well suited to resource-constrained settings requiring both speed and accuracy.}
    \label{fig:imagenet_comparison}
\end{figure}

\cref{fig:imagenet_comparison} compares fast GSPN (Tiny/Small/Base) with leading CNN, Transformer, and SSM architectures on ImageNet-1K, focusing on Top-1 accuracy, throughput (images/second), and parameters at $224^2$. Taking the Tiny model as an example:
\begin{itemize}
    \item \textbf{CNNs:} ConvNeXt-T reaches $82.1\%$ with 29M parameters, 4.5G FLOPs, and $1189$ img/s; ConvNeXt-B reaches $83.8\%$ at 89M/15.4G but only $435$ img/s.
    \item \textbf{Transformers:} DeiT-S has 22M/4.6G, $79.8\%$, $1759$ img/s; Swin-B (88M/15.4G) reaches $83.5\%$ at $458$ img/s.
    \item \textbf{SSMs:} VMamba-T gives $1686$ img/s with 30M/4.9G at $82.6\%$; LocalVMamba-T uses 26M/5.7G for $82.7\%$ but only $394$ img/s.
    \item \textbf{Fast GSPN-T (ours):} $83.0\%$ Top-1 with only 24M parameters and 3.6G FLOPs at $1544$ img/s. Versus DeiT-S, $+3.2\%$ accuracy with 2M more parameters and 1G fewer FLOPs; versus VMamba-T, $+0.4\%$ accuracy with 6M fewer parameters and 1.3G fewer FLOPs at comparable throughput.
\end{itemize}
Fast GSPN thus offers an excellent accuracy/size/efficiency trade-off, with strong throughput for its accuracy class.

\section{Performance with Varying Batch and Channel Dimensions}
\label{app:perf_batch_channel}

\begin{figure}[t]
    \centering
    \includegraphics[width=\textwidth]{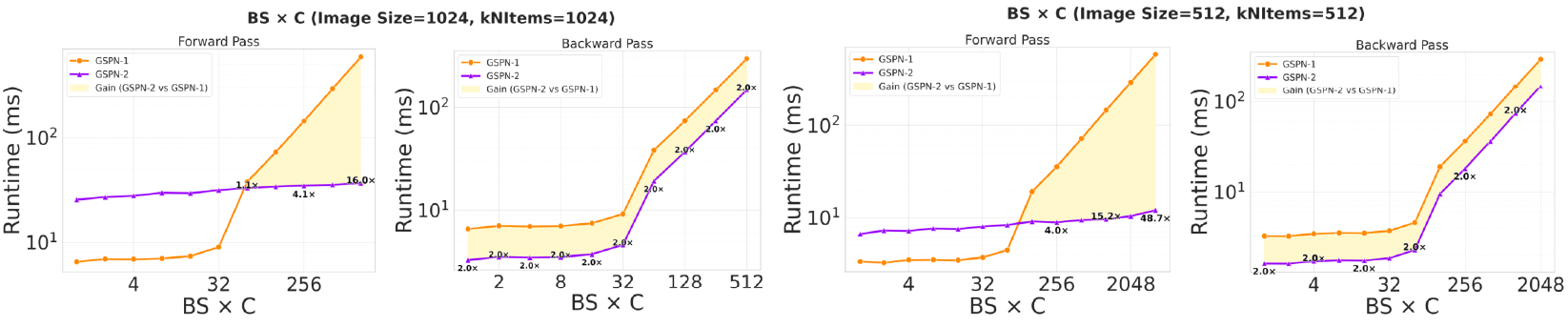}
    \caption{\textbf{Runtime comparison of the original GSPN kernel and the fast GSPN kernel} across different \emph{batch size $\times$ channel} products. Fast GSPN's advantage grows markedly as $B\times C$ increases.}
    \label{fig:runtime_app}
\end{figure}

\cref{fig:runtime} (main paper) shows the fast GSPN kernel's largest speedups arise when batch size or channel count is large. \cref{fig:runtime_app} examines when the full fast GSPN kernel optimizations (including shared-memory caching) begin to dominate, as a function of the $B\times C$ product---a key concern for visual-encoder training and video processing.

\noindent\textbf{Implications for model selection.}
The effectiveness of fast GSPN's most advanced kernel optimizations, such as shared-memory caching of hidden states, is magnified as the aggregate workload ($B\times C$) increases; for very large effective batch sizes (large-scale training or high-throughput video), deploying the fully optimized fast GSPN kernel is critical. Conversely, when $B\times C$ is small, the difference between a lightweight kernel variant and the full fast GSPN kernel is modest, suggesting an adaptive strategy that selects the kernel configuration based on input dimensions and batch size.

\subsection{Optimization Analysis under a Large-Batch Configuration}
\label{app:effablat_large_batch}

\begin{figure}[t]
    \centering
    \includegraphics[width=0.5\textwidth]{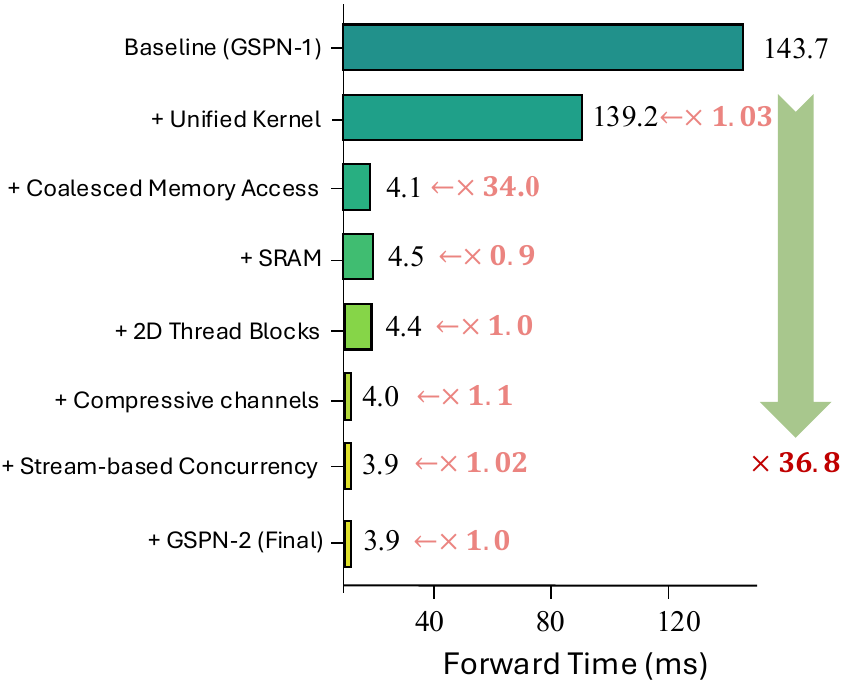}
    \caption{\textbf{Step-by-step CUDA optimization under a large-batch configuration} ($1024\times1024$, batch $256$, $1$ channel). Cumulative reduction in forward time (ms); $36.8\times$ speedup from the original GSPN ($143.7$ ms) to the fast GSPN kernel ($3.9$ ms).}
    \label{fig:journey_app}
\end{figure}

\begin{figure}[t]
    \centering
    \includegraphics[width=0.5\textwidth]{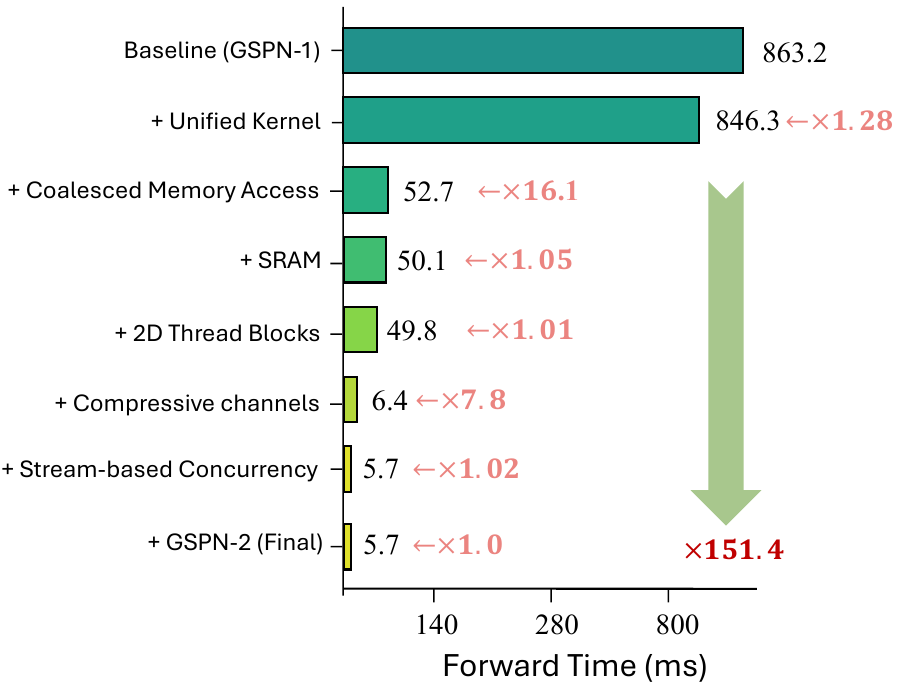}
    \caption{\textbf{Step-by-step CUDA optimization under a large-channel configuration} ($1024\times1024$, batch $1$, $1152$ channels). The \emph{compressive channels} term dominates ($7.8\times$), for an overall $151.4\times$ speedup ($863.2$ ms $\to$ $5.7$ ms).}
    \label{fig:journey_app2}
\end{figure}

While \cref{fig:journey} analyzes a moderate configuration ($1024\times1024$, batch $16$, $8$ channels), \cref{fig:journey_app} examines a high-throughput scenario ($1024\times1024$, batch $256$, $1$ channel), representative of batch video processing or multi-stream serving. The progression: GSPN baseline $143.7$ ms; \emph{unified kernel} $1.03\times$ ($139.2$ ms); \emph{coalesced memory access} $34.0\times$ ($4.1$ ms, even larger than the $23.9\times$ of the 8-channel case, since coalescing is critical at large batch); \emph{shared memory} $0.9\times$ ($4.5$ ms, a slight slowdown because with one channel the L1 cache already captures reuse and explicit SRAM management adds overhead); \emph{2D thread blocks} $1.0\times$ ($4.4$ ms, neutral with a single channel); and \emph{compressive channels} $1.1\times$ ($4.0$ ms, with $3.9$ ms after fine-tuning). The cumulative $36.8\times$ speedup is comparable to the $40.0\times$ of the main configuration, confirming fast GSPN generalizes across workloads while the \emph{relative} importance of each optimization is configuration-dependent. \cref{fig:journey_app2} shows the complementary large-channel scenario ($1024\times1024$, batch $1$, $1152$ channels): here \emph{compressive channels} becomes the dominant optimization, achieving a $7.8\times$ speedup (from $49.8$ ms to $6.4$ ms) by reducing the effective channel dimension $8\times$, for an overall $151.4\times$ speedup ($863.2$ ms $\to$ $5.7$ ms). This validates that channel compression is especially impactful for wide feature maps common in vision transformers and diffusion models.

\section{Text-to-Image Generation}
\label{app:t2i}

\begin{figure}[h]
\begin{minipage}{0.5\textwidth}
\centering
\footnotesize
\captionof{table}{\textbf{Generation on COCO at $1024\times1024$.} Lower FID ($\downarrow$) and higher CLIP-T ($\uparrow$) are better.}
\label{tab:t2i}
\begin{tabular}{l|cc}
\toprule
Model & FID($\downarrow$)    & CLIP-T($\uparrow$) \\
\midrule
SD-v1.5 (baseline)  & 32.71  & 0.290 \\
Mamba~\citep{gu2023mamba} (w/ norm) & 50.30  & 0.263 \\
Mamba2~\citep{dao2024transformers} (w/ norm) & 37.02  & 0.273 \\
Linfusion~\citep{liu2024linfusion} (w/ norm) & 36.33  & 0.285 \\
\midrule
GSPN & 30.86  & 0.307 \\
\textbf{Fast GSPN (Ours)} & 33.21  & 0.286 \\
\bottomrule
\end{tabular}
\end{minipage}%
\begin{minipage}{0.5\textwidth}
    \centering
    \includegraphics[width=0.95\textwidth]{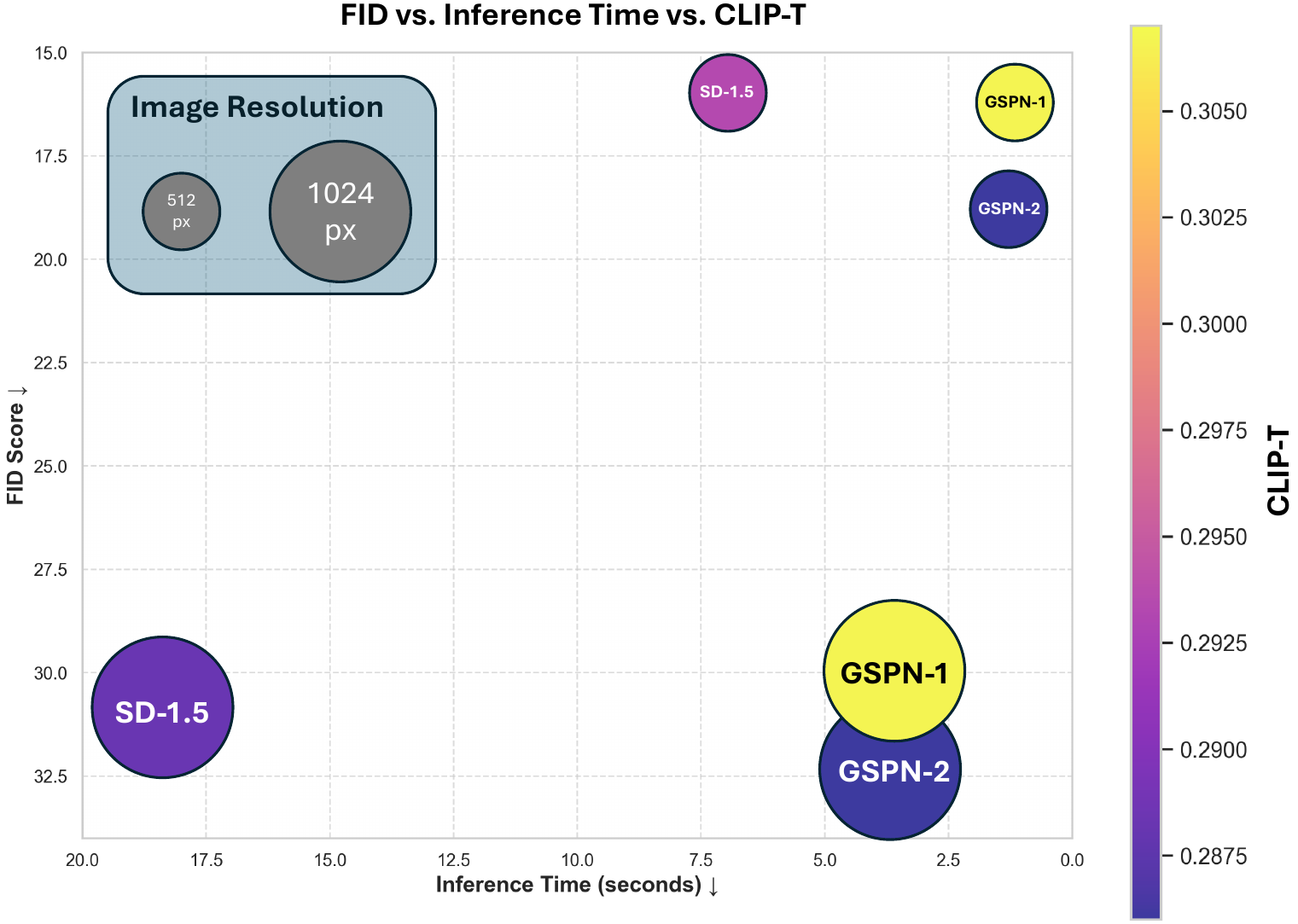}
    \captionof{figure}{\textbf{Fast GSPN vs.\ the original GSPN and baselines.} Fast GSPN achieves a good trade-off between FID, CLIP-T, and inference time.}
    \label{fig:t2i_bubble}
\end{minipage}
\end{figure}

We evaluate fast GSPN on text-to-image generation against the original GSPN and several baselines on COCO at $1024\times1024$ (\cref{tab:t2i}). The baseline is Stable Diffusion v1.5 (SD-v1.5)~\citep{rombach2022high}; we also include Mamba~\citep{gu2023mamba}, Mamba2~\citep{dao2024transformers}, and Linfusion~\citep{liu2024linfusion}, for which text embeddings are treated as part of the visual token sequence.

Fast GSPN attains FID $33.21$ and CLIP-T $0.286$, competitive with and close to SD-v1.5 (FID $32.71$, CLIP-T $0.290$) while inferring faster. Like the original GSPN, fast GSPN adapts to arbitrary resolutions without extra normalization for unseen sizes, since the Stability--Context property ensures stable long-range propagation---an advantage over Mamba and Linfusion, which require resolution-specific normalization.

\section{Compressive Proxy Dimension as Low-Rank Approximation}
\label{app:lowrank}

The compressive proxy strategy (\cref{sec:gspnv2_theory}) projects $\mathbf{X}\in\mathbb{R}^{B\times C\times H\times W}$ to a compressed space $\mathbf{X}_{\text{proxy}}\in\mathbb{R}^{B\times C_{\text{proxy}}\times H\times W}$ ($C_{\text{proxy}}\!\ll\!C$), applies propagation there, and projects back---analogous to low-rank factorization. It reduces the CUDA workload from $k_{\text{chunk}}\times B\times C$ slices to $k_{\text{chunk}}\times B\times C_{\text{proxy}}$, preventing GPU saturation while preserving representational capacity. \cref{tab:proxy_ablation} reports an ablation on $C_{\text{proxy}}$ for fast GSPN-Tiny on ImageNet-1K.

\begin{table}[h]
\centering
\caption{\textbf{Ablation on proxy dimension $C_{\text{proxy}}$} (Fast GSPN-Tiny, ImageNet-1K).}
\label{tab:proxy_ablation}
\begin{tabular}{c|c|c}
\toprule
$C_{\text{proxy}}$ & Accuracy (\%) & Throughput (img/s) \\
\midrule
2  & 83.0 & 1544 \\
4  & 83.0 & 1492 \\
8  & 83.0 & 1387 \\
16 & 82.9 & 1293 \\
32 & 82.8 & 1106 \\
\bottomrule
\end{tabular}
\end{table}

Accuracy degrades minimally ($0.2\%$ from $C_{\text{proxy}}{=}32$ to $C_{\text{proxy}}{=}2$) while throughput improves $1.4\times$. The aggressive $48{:}1$ compression at $C_{\text{proxy}}{=}2$ shows that GSPN propagation operates effectively in low-dimensional spaces, as spatial dependencies dominate over channel-wise ones.

\section{C-GSPN Implementation Details}
\label{app:cgspn_impl}

\subsection{Pretraining}
Before end-to-end distillation, we run a lightweight pretraining stage to stabilize optimization and provide a strong initialization. We train on $5$M image--text pairs sampled from DataComp~\citep{gadre2023datacomp} using AdamW~\citep{loshchilov2017decoupled} with learning rate $4\times10^{-5}$, global batch size $1024$, $300$ warmup steps, and a linear decay schedule. Omitting this step leads to unstable early-epoch training and consistently lower downstream performance.

\subsection{End-to-End Distillation Training}
For full-scale training we distill C-GSPN on $600$M curated image--text pairs from DataComp, aligning the student to its teacher (OpenCLIP SO/14) through the staged supervision of \cref{sec:distill}. We adopt a sparse distillation strategy, supervising every ninth teacher block, and use AdamW with learning rate $4\times10^{-4}$, global batch size $8192$, and a cosine schedule with $10{,}000$ warmup steps.

\subsection{Loss Composition and Balancing}
The total distillation loss combines the two taps per block---post-propagation (PP) and post-block (PB):
\begin{equation}
\mathcal{L} = \alpha\,\mathcal{L}_{\text{PP}} + \beta\,\mathcal{L}_{\text{PB}},
\end{equation}
with each term combining MSE feature alignment and KL distribution matching (\cref{eq:distill_losses}). We set $\alpha=\beta=0.5$ to balance PP and PB supervision, ensuring the propagation sublayer is constrained directly without being overshadowed by block-level matching, and $\lambda_1=\lambda_2=7/3$ to balance MSE and KL. A lightweight 2-layer MLP adaptor is inserted before each tap (\cref{sec:distill}); the overall two-stage scheme is illustrated in \cref{fig:distill-loss} of the main text.

\begin{figure}[h]
    \centering
    \includegraphics[width=0.95\columnwidth]{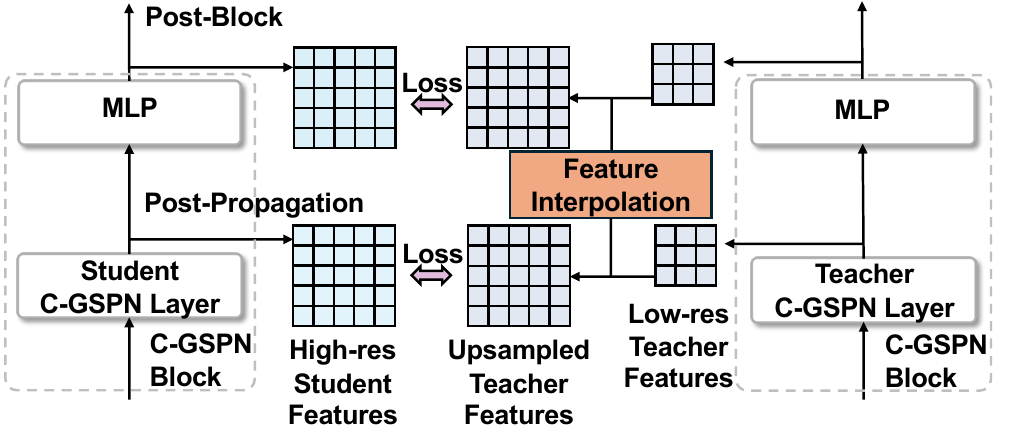}
    \caption{\textbf{High-resolution encoder distillation.} A frozen low-resolution teacher supervises a higher-resolution student via upsampled features at two taps (post-propagation and post-block), with feature interpolation bridging resolutions, applied progressively in a resolution curriculum. See \cref{sec:highres-transfer}.}
    \label{fig:highres-transfer}
\end{figure}

\subsection{Stability Practices}
Sublayer-wise pretraining (Stage 1) provides consistent per-sublayer signals before end-to-end optimization (Stage 2). In ablations, removing either the adaptors or Stage 1 degrades stability and final accuracy.

\subsection{Replacing $1/9$ of GSPN Blocks with Attention}
For the hybrid of \cref{sec:ablation_study}, we start from the 27-block GSPN backbone and evenly interleave attention blocks through depth: every ninth GSPN block is replaced by a standard multi-head self-attention block with the same embedding dimension and MLP, yielding a $3/27\approx1/9$ attention ratio. This keeps depth and parameterization comparable while injecting sparse pairwise mixing at regular intervals, and is the variant reported as the ``$1/9$-attention hybrid''.

\section{Additional C-GSPN Results}
\label{app:more_latency}

\subsection{Multi-Resolution Distillation}
We train a single C-GSPN model operating across multiple input resolutions without special positional embeddings, supervised by a low-resolution teacher. \cref{tab:multi-resolution-distillation} shows the student maintains comparable ADE20K performance across $378$, $448$, and $518$, indicating effective cross-scale transfer.

\begin{table}[h]
    \centering
    \resizebox{0.8\columnwidth}{!}{%
    \begin{tabular}{l c c c c}
        \toprule
        Dataset & 378-teacher & 378-multires & 448-multires & 518-multires \\
        \midrule
        ADE20K &  46.0  & 45.8  & 45.8  & 45.9  \\
        \bottomrule
    \end{tabular}
    }
    \caption{\textbf{Multi-resolution distillation on ADE20K (mIoU).} A single multi-resolution student matches the single-resolution baseline.}
    \label{tab:multi-resolution-distillation}
\end{table}

\subsection{More Latency and Throughput Analysis}
To complement \cref{sec:performance}, \cref{tab:throughput} reports end-to-end throughput at batch size $32$ across three resolutions, and \cref{tab:latency_breakdown_full} decomposes latency into sublayer, layer, and full-block timings. The propagation sublayer remains stable while attention baselines quickly become memory- or latency-bound, with many configurations running out of memory (OOM) at high resolution.

\begin{table*}[h]
    \centering
    \begin{tabular}{l c c c c c c}
        \toprule
        & \multicolumn{6}{c}{\textbf{Model Throughput (img/s)}} \\
        \cmidrule(lr){2-7}
        & \multicolumn{2}{c}{\textbf{518}} & \multicolumn{2}{c}{\textbf{1554}} & \multicolumn{2}{c}{\textbf{2590}} \\
        \cmidrule(lr){2-3} \cmidrule(lr){4-5} \cmidrule(lr){6-7}
        \textbf{Method} & \textbf{img/s} & \textbf{$\times$} & \textbf{img/s} & \textbf{$\times$} & \textbf{img/s} & \textbf{$\times$} \\
        \midrule
        Attention      & 49.42 & \cellcolor[HTML]{F1FAF1}2.28$\times$ & OOM & \cellcolor[HTML]{F1FAF1}OOM & OOM & \cellcolor[HTML]{F1FAF1}OOM \\
        FlashAttention & 104.06 & \cellcolor[HTML]{F1FAF1}1.08$\times$ & 4.78 & \cellcolor[HTML]{F1FAF1}2.58$\times$ & 0.78 & \cellcolor[HTML]{F1FAF1}5.32$\times$ \\
        GSPN           & 25.53 & \cellcolor[HTML]{F1FAF1}4.42$\times$ & OOM & \cellcolor[HTML]{F1FAF1}OOM & OOM & \cellcolor[HTML]{F1FAF1}OOM \\
        \rowcolor[HTML]{D0E8D0}\textbf{C-GSPN (ours)}  & \textbf{112.91} & \textbf{1$\times$} & \textbf{12.35} & \textbf{1$\times$} & \textbf{4.15} & \textbf{1$\times$} \\
        \bottomrule
    \end{tabular}
    \caption{\textbf{Model throughput} (img/s) at batch size $32$ across three resolutions. {\setlength{\fboxsep}{1pt}\colorbox[HTML]{F1FAF1}{\strut Shaded columns}} report the multiplicative gap vs.\ ours. C-GSPN maintains practical throughput at all resolutions while competing methods run out of memory or degrade sharply.}
    \label{tab:throughput}
\end{table*}

\begin{table*}[h]
    \centering
    \resizebox{\textwidth}{!}{%
    \begin{tabular}{l c c c c c c c c c c c c c c c c c c}
        \toprule
        & \multicolumn{6}{c}{\textbf{Sublayer}} & \multicolumn{6}{c}{\textbf{Layer}} & \multicolumn{6}{c}{\textbf{Block}} \\
        \cmidrule(lr){2-7} \cmidrule(lr){8-13} \cmidrule(lr){14-19}
        & \multicolumn{2}{c}{\textbf{518}} & \multicolumn{2}{c}{\textbf{1554}} & \multicolumn{2}{c}{\textbf{2590}} & \multicolumn{2}{c}{\textbf{518}} & \multicolumn{2}{c}{\textbf{1554}} & \multicolumn{2}{c}{\textbf{2590}} & \multicolumn{2}{c}{\textbf{518}} & \multicolumn{2}{c}{\textbf{1554}} & \multicolumn{2}{c}{\textbf{2590}} \\
        \cmidrule(lr){2-3} \cmidrule(lr){4-5} \cmidrule(lr){6-7} \cmidrule(lr){8-9} \cmidrule(lr){10-11} \cmidrule(lr){12-13} \cmidrule(lr){14-15} \cmidrule(lr){16-17} \cmidrule(lr){18-19}
        \textbf{Method} & \textbf{ms} & \textbf{$\times$} & \textbf{ms} & \textbf{$\times$} & \textbf{ms} & \textbf{$\times$} & \textbf{ms} & \textbf{$\times$} & \textbf{ms} & \textbf{$\times$} & \textbf{ms} & \textbf{$\times$} & \textbf{ms} & \textbf{$\times$} & \textbf{ms} & \textbf{$\times$} & \textbf{ms} & \textbf{$\times$} \\
        \midrule
        Attention      & 14.905 & \cellcolor[HTML]{F1FAF1}169.38$\times$ & OOM & \cellcolor[HTML]{F1FAF1}OOM & OOM & \cellcolor[HTML]{F1FAF1}OOM & 17.919 & \cellcolor[HTML]{F1FAF1}4.06$\times$ & OOM & \cellcolor[HTML]{F1FAF1}OOM & OOM & \cellcolor[HTML]{F1FAF1}OOM & 22.900 & \cellcolor[HTML]{F1FAF1}2.44$\times$ & OOM & \cellcolor[HTML]{F1FAF1}OOM & OOM & \cellcolor[HTML]{F1FAF1}OOM \\
        FlashAttention & 2.344 & \cellcolor[HTML]{F1FAF1}26.64$\times$ & 164.696 & \cellcolor[HTML]{F1FAF1}550.82$\times$ & 1259.135 & \cellcolor[HTML]{F1FAF1}1949.28$\times$ & 5.160 & \cellcolor[HTML]{F1FAF1}1.17$\times$ & 190.124 & \cellcolor[HTML]{F1FAF1}5.07$\times$ & 1336.021 & \cellcolor[HTML]{F1FAF1}11.70$\times$ & 10.141 & \cellcolor[HTML]{F1FAF1}1.08$\times$ & 235.372 & \cellcolor[HTML]{F1FAF1}2.85$\times$ & 1466.011 & \cellcolor[HTML]{F1FAF1}6.00$\times$ \\
        GSPN           & 3.415 & \cellcolor[HTML]{F1FAF1}38.81$\times$ & OOM & \cellcolor[HTML]{F1FAF1}OOM & OOM & \cellcolor[HTML]{F1FAF1}OOM & 30.391 & \cellcolor[HTML]{F1FAF1}6.89$\times$ & OOM & \cellcolor[HTML]{F1FAF1}OOM & OOM & \cellcolor[HTML]{F1FAF1}OOM & 35.372 & \cellcolor[HTML]{F1FAF1}3.77$\times$ & OOM & \cellcolor[HTML]{F1FAF1}OOM & OOM & \cellcolor[HTML]{F1FAF1}OOM \\
        \rowcolor[HTML]{D0E8D0}\textbf{C-GSPN (ours)}  & \textbf{0.088} & \textbf{1$\times$} & \textbf{0.299} & \textbf{1$\times$} & \textbf{0.646} & \textbf{1$\times$} & \textbf{4.413} & \textbf{1$\times$} & \textbf{37.472} & \textbf{1$\times$} & \textbf{114.170} & \textbf{1$\times$} & \textbf{9.394} & \textbf{1$\times$} & \textbf{82.720} & \textbf{1$\times$} & \textbf{244.160} & \textbf{1$\times$} \\
        \bottomrule
    \end{tabular}%
    }
    \caption{\textbf{Latency breakdown across architectural levels} (batch size $32$). Sublayer: core 2D propagation vs.\ attention sublayer; Layer: full layer; Block: complete transformer block (ms). {\setlength{\fboxsep}{1pt}\colorbox[HTML]{F1FAF1}{\strut Shaded columns}} report the multiplicative gap vs.\ ours. C-GSPN's sublayer is up to $1949\times$ faster than FlashAttention at $2590$ and $6\times$ faster on end-to-end block latency.}
    \label{tab:latency_breakdown_full}
\end{table*}

\subsection{Downstream Vision-Language Use}
We additionally pair the C-GSPN encoder with a Qwen2.5-3B language model and evaluate on VQA-style tasks (\cref{tab:vqa}); C-GSPN closely matches the OpenCLIP encoder while offering substantially lower high-resolution latency.

\begin{table}[h!]
    \centering
    \resizebox{0.5\columnwidth}{!}{%
    \begin{tabular}{l c  c c}
        \toprule
        \textbf{VQA tasks} & \textbf{Seed-Img} & \textbf{VizWiz} & \textbf{MMMU} \\
        \midrule
        \rowcolor[HTML]{F1F5FB} \textbf{OpenCLIP+Qwen2.5-3B} & 72.9  &  55.5   & 24.4 \\
        \rowcolor[HTML]{D0E8D0}\textbf{C-GSPN+Qwen2.5-3B}& 72.1  & 55.8    & 24.2  \\
        \bottomrule
    \end{tabular}}
    \caption{\textbf{Vision-language evaluation.} C-GSPN as the vision encoder paired with Qwen2.5-3B.}
    \label{tab:vqa}
\end{table}

\clearpage
\begingroup
\raggedright
\sloppy
\makeatletter
\renewcommand{\bibsection}{%
  \par\noindent{\headingfont\refname}\par\vspace{5pt}%
}
\setlength{\bibhang}{0pt}
\renewcommand\@bibsetup[1]{%
  \setlength{\leftmargin}{0pt}%
  \setlength{\itemindent}{0pt}%
  \setlength{\labelwidth}{0pt}%
  \setlength{\labelsep}{0pt}%
  \setlength{\listparindent}{0pt}%
  \setlength{\itemsep}{\bibsep}%
}
\makeatother
\setlength{\emergencystretch}{3em}
\Urlmuskip=0mu plus 1mu\relax
\bibliographystyle{plainnat}
\bibliography{main}
\endgroup

\end{document}
